\documentclass[review]{elsarticle}

\usepackage{lineno,hyperref}
\usepackage{subfigure}
\usepackage{booktabs}
\usepackage{multirow}
\usepackage{graphicx}
\usepackage{pdfpages}
\usepackage{color}
\usepackage{amssymb}
\usepackage{pifont}
\usepackage{amsmath}
\newcommand{\xmark}{\ding{55}}
\newcommand{\cmark}{\ding{51}}
\usepackage{makecell}
\journal{Journal of \LaTeX\ Templates}

\begin{document}

\begin{frontmatter}

\title{Non-uniform Motion Deblurring with Blurry Component Divided Guidance}


\author[mymainaddress]{Pei Wang}
\ead{wangpei23@mail.nwpu.edu.cn}

\author[mymainaddress]{Wei Sun}
\author[ade]{Qingsen Yan}
\author[mymainaddress]{Axi Niu}
\author[mymainaddress]{Rui Li}
\author[mymainaddress]{Yu Zhu\corref{mycorrespondingauthor}}\ead{yuzhu@nwpu.edu.cn}
\author[mysecondaryaddress]{Jinqiu Sun}
\author[mymainaddress]{Yanning Zhang}

\cortext[mycorrespondingauthor]{Corresponding author}

\address[mymainaddress]{School of Computer Science, Northwestern Polytechnical University, Xi'an 710072, China}
\address[mysecondaryaddress]{School of Astronautics, Northwestern Polytechnical University, Xi'an 710072, China}
\address[ade]{School of Computer Science, The University of Adelaide, Adelaide 5005, Australia}
\address[mythirdaryaddress]{National Engineering Laboratory for Integrated Aero-Space-Ground-Ocean Big Data Application Technology, Xi'an 710072, China}

\begin{abstract}
Blind image deblurring is a fundamental and challenging computer vision problem, which aims to recover both the blur kernel and the latent sharp image from only a blurry observation. Despite the superiority of deep learning methods in image deblurring have displayed, there still exists major challenge with various non-uniform motion blur. Previous methods simply take all the image features as the input to the decoder, which handles different degrees (\textit{e.g.} large blur, small blur) simultaneously, leading to challenges for sharp image generation. To tackle the above problems, we present a deep two-branch network to deal with blurry images via a component divided module, which divides an image into two components based on the representation of blurry degree. Specifically, two component attentive blocks are employed to learn attention maps to exploit useful deblurring feature representations on both large and small blurry regions. Then, the blur-aware features are fed into two-branch reconstruction decoders respectively. In addition, a new feature fusion mechanism, orientation-based feature fusion, is proposed to merge sharp features of the two branches. Both qualitative and quantitative experimental results show that our method performs favorably against the state-of-the-art approaches.

\end{abstract}


\end{frontmatter}


\section{Introduction}
\label{Introduction}
Image deblurring has been an important task in computer vision and image processing for a long time. Given a blurry image, the goal of deblurring is to recover a sharp latent image with necessary edge structures and details. It is a serious ill-posed problem, especially with the non-uniform blur, in which the blur kernel is unknown and the motion is more complicated. The non-uniform deblurring remains a challenging task in computer vision even though several effective methods have been developed.

Traditional learning-free methods applied various constraints to model characteristics of blur \cite{Xu2010Two,Cho2009Fast,gong2017mpgl,hsieh109blind,peng2020joint,li2019new} and utilized different natural image priors \cite{levin2011understanding,pan2014deblurring,gong2016blind,krishnan2009fast} to regularize the solution space. However, these works relied on several model assumptions and are difficult to deal with real blurry images with more realistic motions. 
With the recent development on deep learning, deblurring methods based convolutional neural network (CNN) have drawn more attention and achieved impressive results \cite{sun2015learning,nah2017deep,li2018learning,tao2018scale-recurrent,zhang2019deep,suin2020spatially,purohit2020region,kaufman2020deblurring}. Different from those traditional methods, the approaches based deep learning focus on not only the blur kernel estimation \cite{sun2015learning} and image deconvolution \cite{kaufman2020deblurring}, but also the end-to-end mapping functions \cite{suin2020spatially,purohit2020region} from blurry images to sharp images. 

Due to the advantages of the end-to-end learning method, which always ignores the different stages of traditional method and replaces them with a single neural network, the recent methods mainly adopt an end-to-end learning approach to non-uniform deblurring. Although the existing methods can handle blurry images and have made significant progress, there are still many specific samples caused by non-uniform blur that are difficult to process. To solve the deblurring task from the source, we begin by asking a question: what are the properties of the different degrees of blurry image?

\begin{figure}[htbp]
\centering
\includegraphics[width=4.7in]{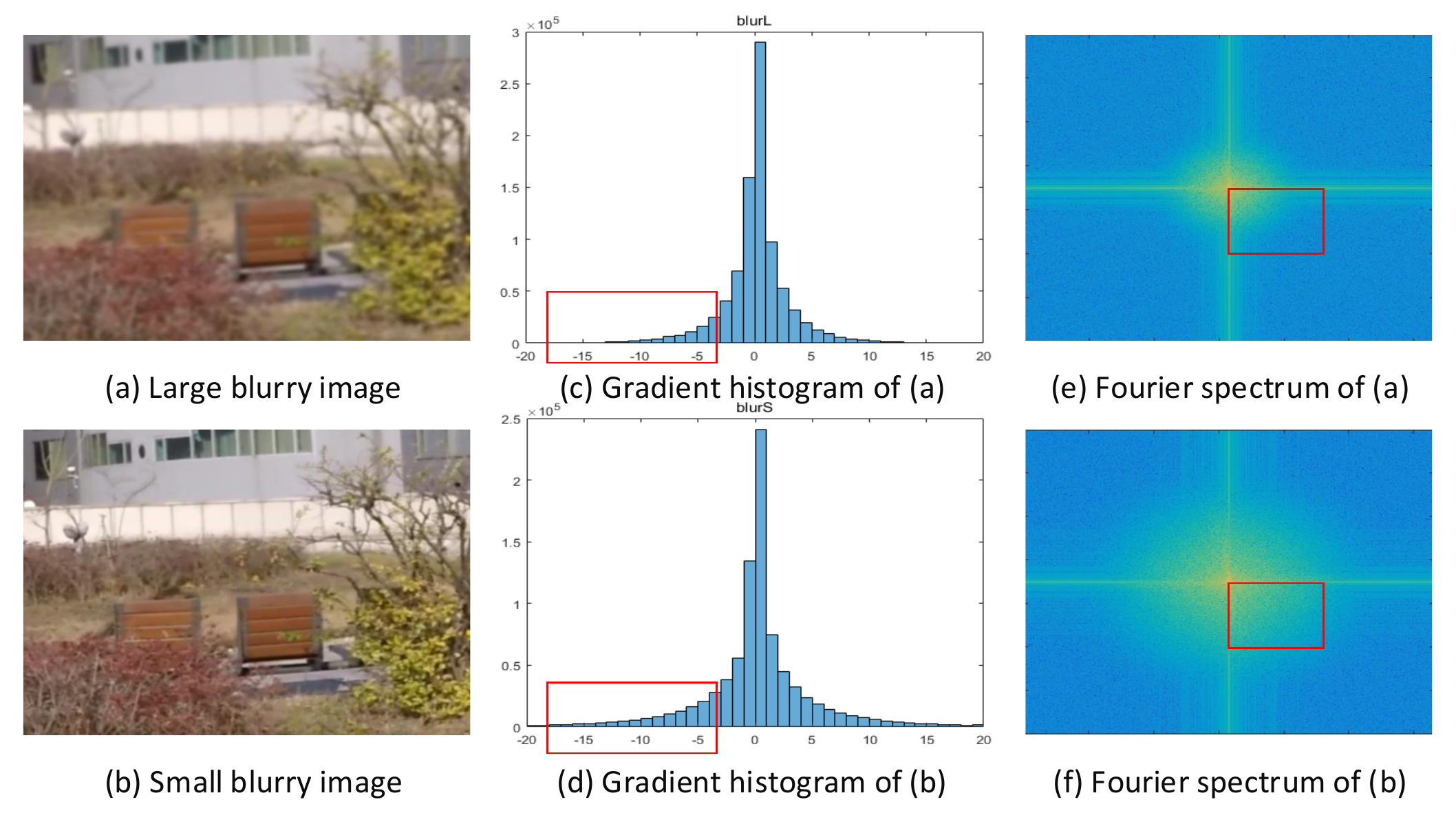}
\caption{The differences between the large and small blurry images on gradient histogram and Fourier spectrum.}
\label{figure_fft}
\end{figure}

We illustrate a large blurry image and a small blurry image in Fig.\ref{figure_fft}, as well as their gradient histograms and Fourier spectrum. Generally speaking, the gradient of a sharp image obeys a heavy-tailed distribution \cite{Fergus2006}, which can be used to guide the deblurring. We can observe two characters from Fig.\ref{figure_fft}. First, the tail of gradient histogram of the small blurry image is longer than that of the large blurry image, and the pixel number of zero-grad of the large blurry image is higher (See Fig.\ref{figure_fft}(c) and (d)).
Second, the Fourier spectrum of the small blurry image , which represents the frequency information, has strong response on high-frequency areas (See Fig.\ref{figure_fft}(e) and (f)). We have reason to believe that the characters (\emph{e.g.} visual effect, gradient domain, Fourier domain) of the image depend on the blur degrees.
Thus, a natural approach is to handle various blurs, separately. 

\begin{figure}[htbp]
\centering
\includegraphics[width=4.7in]{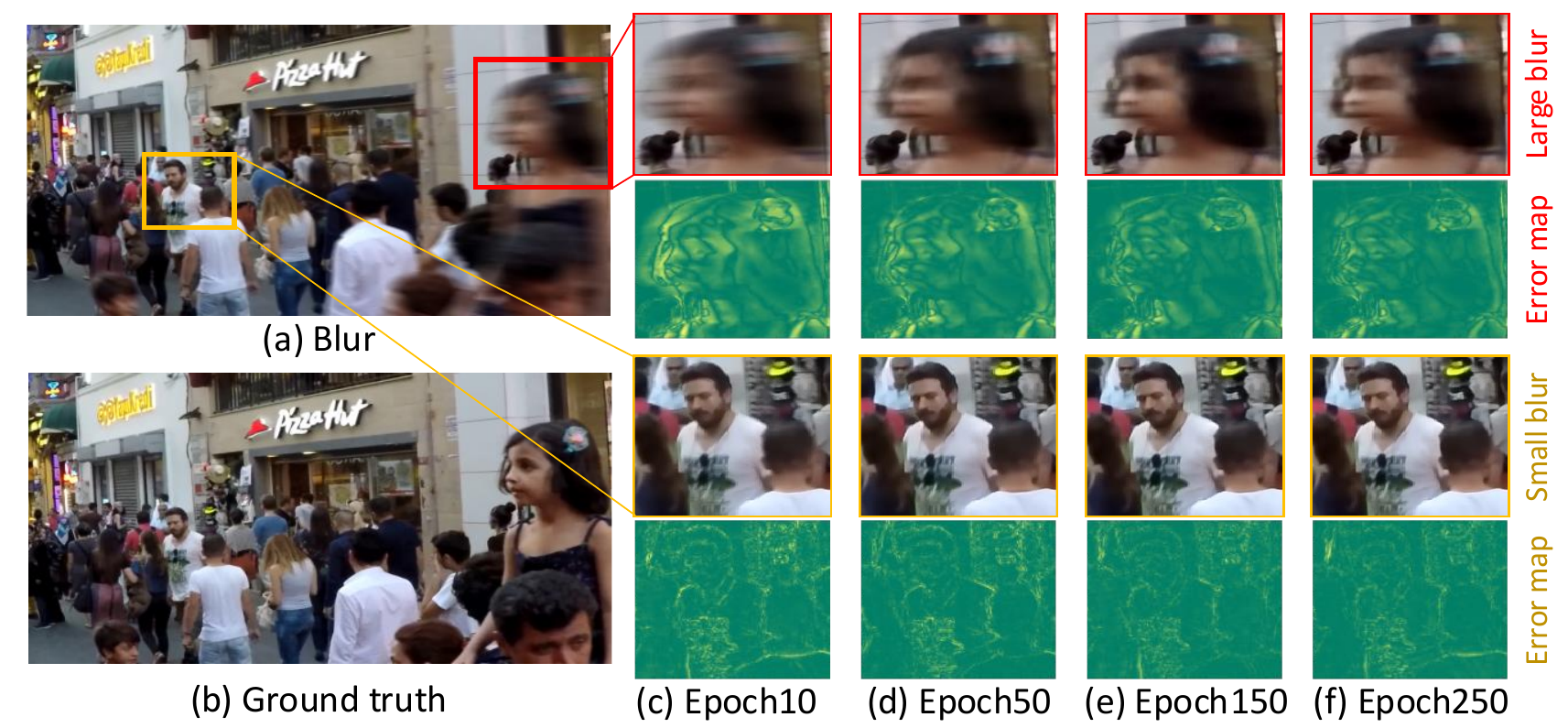}
\caption{The changes and differences of large and small blurry patches from one blurry image during the DMPHN \cite{zhang2019deep} training strategy. (a) and (b) are blurry and ground truth images. From top-right to bottom-right: top two rows: large blurry patch's recovery results and the error maps. Bottom two rows: small blurry patch's results and the error maps which were calculated by ground truth and the deblurred results of each epoch.}
\label{different_regions}
\end{figure}

Unfortunately, most deep-learning approaches learn a non-uniform deblur model under a unified training strategy, which treats the various blurry regions as an entirety to reduce the deblurring solution space, such as DMPHN \cite{zhang2019deep} which attempts to use a multi-patch framework with a unified training strategy to solve the non-uniform blur. The training stage of DMPHN is shown in Fig.\ref{different_regions}. Two patches (marked by red and yellow box) with large and small blurry degree are cropped from one blurry image (a). Under the unified training manner, the large and small blurry patches exhibit a different recovered trend (See (c)-(f)). With the training step goes on, the large blurry patch recovers obviously, however, it's still hard to recovery the latent sharp patch compared to the ground truth (b). For the small blurry patch, due to the relatively light blur degree, after 50 epochs, the small blurry patch changed barely (Bottom two rows: (c)-(f)). Although the recovered details in (f) are almost close to its ground truth, there still exists a gap between them. The error maps which shown in color-map clearly demonstrated the statements.   

Since the recovered performances of the large blurry regions and the small blurry regions are different as the number of training increases (see in Fig.\ref{different_regions}), the latter will be affected by recovery process of the large blur. As the training stage continues, the model has to learn both large and small deblurring parameters in a unified manner, which will eventually cause a sub-optimal solution. Intuitively, the models learned by traditional training strategy prefer addressing regions with large blur, but usually fail to infer realistic details of seemingly blurry regions. In summary, it's difficult to improve the whole image's quality effectively under the unified training strategy. 

Based on the illustrations motioned above, this paper pays more attention to different blurry regions' processing, especially the large blurry regions and small blurry regions. Fig.\ref{overview_en_de} shows the flowchart of our method. We propose a component-divided module with two-branch training strategy to deal with the non-uniform blur with different blur degrees. Under the proposed blurry component divided guidance, the model can handle large and small blurry regions separately and effectively.

Specifically, we first develop a residual dense encoder to extract more deeper features. Then the adaptive component divided attention module is designed for selecting beneficial features of large and small blurry regions to deblurring. The blur-aware features are fed into two-branch decoders, each branch focuses on learning one of the blurry degree components. In this way, each branch is trained separately to deal with the degree of blurry it needs to process. Finally, the results of two branches are aggregated by an orientation based fusion module to generate the final deblurring reconstruction image. The proposed adaptive component divided guidance network under a two-branch training strategy for non-uniform motion deblurring, which explores the motion deblurring by explicitly disentangling the blur of large and small motion. 

The main contributions are summarized as follows:
\begin{itemize}
\item We propose a novel efficient two-branch training strategy to handle the non-uniform blur, which produces the sharp images relying on the large and small blur removal branches, respectively.
\item An adaptive component divided attention module is proposed that can adaptively divide blurry image into large and small blurry regions according to the component of the blurry image.
\item To merge the two branches' deblurred features effectively, an orientation based fusion module is proposed.
\item We quantitatively and qualitatively evaluate our network on benchmark datasets and real-world blurry images, which has shown that our method achieves remarkable performance, but the parameters and computation are smaller than other algorithms compared to previous state-of-the-art methods. 
\end{itemize}

The remainder of this paper is organized as follows. In Section \ref{Related work}, some previous related works are simply shown. Section \ref{proposed method} presents the proposed model and optimization procedure. Section \ref{Experiments} presents the experimental results and analysis. Finally, we conclude our work in Section \ref{conclusion}.                                                                                       
\section{Related work}
\label{Related work}
A blurry image can be modeled using following equation:
\begin{equation}
B = I \otimes K + n,
\end{equation}
where $\otimes$ represents the convolution operator, $B$ is the blurry image, $K$ is the blur kernel, $I$ is the latent sharp image and $n$ is the additive noise respectively, where n always be assumed as Gaussian noise. Under the non-uniform blind deblurring conditions, the $I$, $K$ and $N$ are all unknown, and the $K$ is spatially variant.

\subsection{Non-uniform Image Deblurring} 
The task of image deblurring has been developed for a long time and has undergone a transformation from traditional methods to deep learning methods. 
Conventional image deblurring methods \cite{peng2020joint,li2019new,levin2011understanding,pan2014deblurring,gong2016blind,krishnan2009fast,chen2008robust} fail to remove non-uniform motion blur due to the use of spatially-invariant kernel and assumptions of Gaussian additive noise in images. Moreover, the traditional methods are mostly time-consuming and computationally expensive, especially when there is a nested multi-layer iteration process, it will lead to longer processing time, which cannot satisfy the ever-growing needs for real-time deblurring. 

Recently, deep learning methods have been used on non-uniform deblurring to deal with the complex motion blur in a time-efficient manner. Sun \textit{et.al.} \cite{sun2015learning} and Gong \textit{et.al.} \cite{gong2017motion} used a CNN to estimate the kernel of different blurry patches, and deconvolved a clear image similar to the traditional method. Some methods \cite{nah2017deep,tao2018scale-recurrent,gao2019dynamic} found that the same blurry image patch has different blur representations under different scales, and proposed a multi-scale deblurring network framework to process images with different image size. The deconvolution/upsampling operations in the coarse-to-fine scheme result in expensive runtime and simply increase the model depth with finer-scale levels cannot improve the quality of deblurring obviously, Zhang \textit{et.al.} \cite{zhang2019deep} and Suin \textit{et.al.} \cite{suin2020spatially} proposed a multi-patch network to deal with blurry images via a fine-to-coarse hierarchical representation. Since Generative Adversarial Network (GAN) has been proposed many years, it has been widely used in computer vision task especially in image generation, which can help these tasks to produce more realistic results. Kupyn \textit{et.al.} \cite{kupyn2018deblurgan,kupyn2019deblurgan} designed GAN-based deblurring network that presented a conditional GAN which produces high-quality and more realistic deblurred images via Wasserstein loss. 

So far, Suin \textit{et.al.} \cite{suin2020spatially} and Purohit \textit{et.al.} \cite{purohit2020region} achieve the best performance which based on some complex and computational modules, such as DenseNet. By the way, the model \cite{purohit2020region} has been pretrained on ImageNet is used as the encoder of the whole network also performs well. Sim \textit{et.al.} \cite{sim2019deep} developed a deblurring network based on per-pixel adaptive kernels with residual down-up and up-down modules which makes the excellent performance. Unlike the previous methods that usually attempted to learn a deblurring model under a unified training strategy which ignored the difference among blurry image regions, we propose a two-branch framework under a non-unified training strategy to deal with the large and small degrees blurry regions of input blurry image, respectively. 

\subsection{Attention Mechanism}
Attention mechanism \cite{Bahdanau2014,Vaswani2017} has been used in several computer vision task, such as image style transfer. Chen and Mejjati \textit{et al.} \cite{ChenXYT18,MejjatiRTCK18} used an attention map to guide the network to learn the interest regions of the input and transfer these regions to another domain style, and the most common example is converting a horse of the image into a zebra. Recently, a few methods attempted to adopt the attention mechanism in image deblurring problem, such as self-attention \cite{purohit2020region}. However, the using of the attention mechanism is rare in the field of deblurring. It may be due to the special problems of image blur: 1) For blur, which parts of the image should be paid attention to is not clear enough, 2) It is difficult to directly use the attention map of the whole image to constrain the network to obtain deblurring results. Recently, Shen \textit{et al.} \cite{ShenWLSLX019} proposed an human-aware attention module, which explicitly encoded foreground human information by learning a soft human-mask, and developed foreground and background two decoders branch to focus on their specific domains and suppresses irrelevant information. \cite{suin2020spatially,purohit2020region} both used self-attention as an important part of the network, which played a core role in image deblurring. Zhang \textit{et.al.} \cite{zhang2018image} proposed a very deep residual channel attention network (RCAN) for high-precision image super resolution. \cite{woo2018cbam} designed a convolutional block attention module (CBAM) which combined the spatial and channel attention. Compared with SENet \cite{hu2019senet} who only focused on channels attention mechanism, CBAM can achieve the better results. Inspired by \cite{ShenWLSLX019,zhang2018image,woo2018cbam}, in this paper, we proposed a advanced attention module named adaptive component divided attention (ACDA) which can divide the blurry regions into large and small parts.  

\subsection{Deformable Convolution}
Deformable convolutional networks was firstly proposed by Dai \textit{et.al.} \cite{Dai2017Deformable}. Because the geometry in the modules used to construct regular convolutional neural networks (CNNs) is fixed, their ability to model geometric transformations is inherently limited. Deformable adjusts the displacement of the spatially sampled position information in the module. The displacement can be learned in the target task and does not require additional supervision signals. Since it can improve the capability of regular convolutions and enlarge the receptive, it's widely used in various tasks such as video object detection, semantic segmentation \cite{Dai2017Deformable} and video super-resolution \cite{WangCYDL19}. The sampling position of deformable convolution is more in line with the shape and size of the image's object itself, while the regular standard convolution cannot. Inspired by this property, the deformable convolution can be great useful to process the image object's structure and shape caused by blur. Following \cite{purohit2020region}, we leverage the deformable convolution to realize our two reconstruction decoders which can effectively solve the different degrees blurry regions from one blurry image in various ways.

\subsection{Feature Fusion}
In many deep-learning-based image processing methods, feature fusion techniques, which also named feature alignment, play an important role in combining the different features of various scales or sizes. The simplest operator is concatenation or addition that directly aggregates the different features from the different branch. Some approaches \cite{li2018learning,dogan2019exemplar} adopt the concatenation-based fusion, which doesn't consider the illumination difference and spatial variation. \cite{Li_2020_CVPR} proposed an adaptive spatial feature fusion layer to incorporate features in an adaptive and progressive manner. \cite{qiu2019embedded} proposed a recurrent fusion technique to single image super resolution that stabilizes the feature flow and the gradient flow in training and encourages a faster convergence rate of training. Due to the strongly directionality of blur, in this work, we design an orientation-based fusion module to effectively combine the two-branch reconstruction results.

\section{Blurry Component Divided Guided Network}
\label{proposed method}
\begin{figure*}[htbp]
\centering
\includegraphics[width=4.8in]{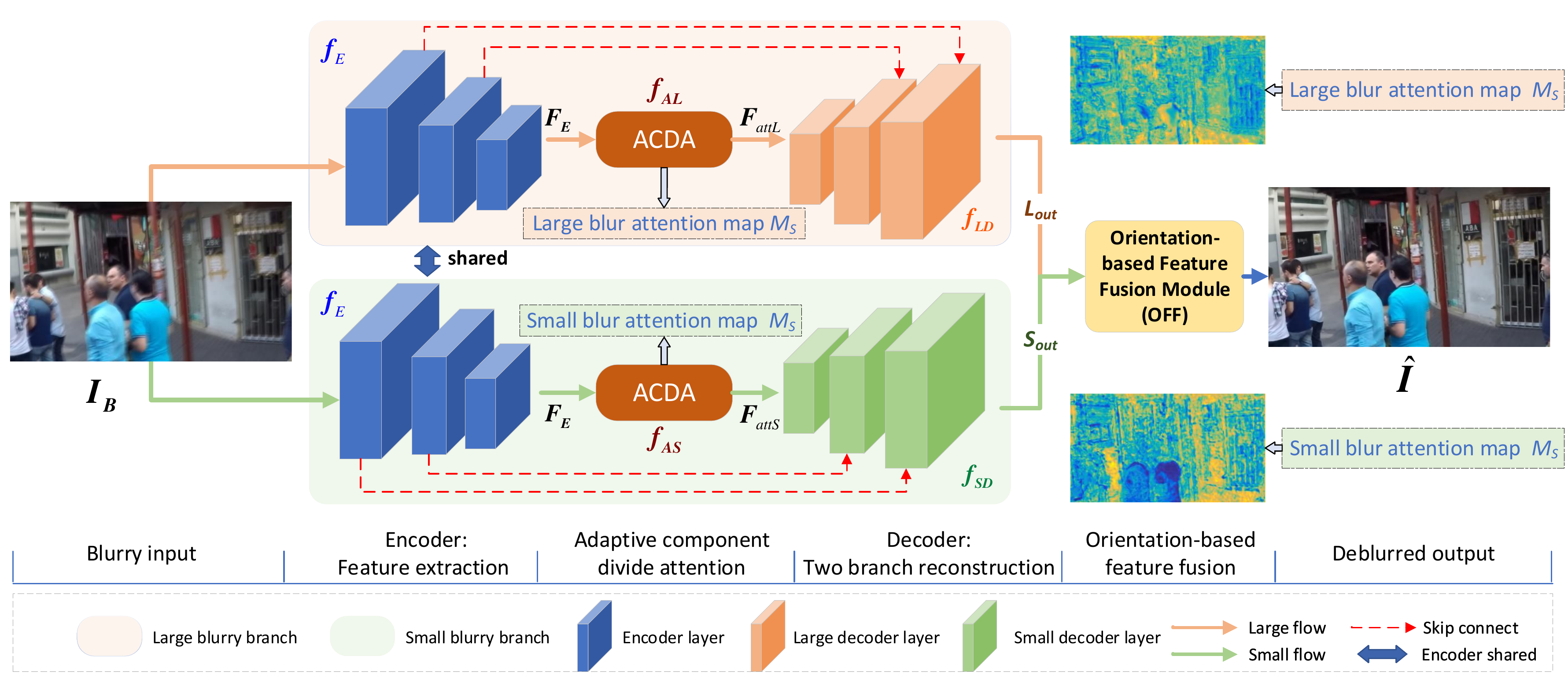}
\caption{The overview of the proposed blurry component divided guided network. We use an encoder-decoder architecture to build our network. The ACDA module is connected with the encoder and decoder, finally followed by an orientation based fusion module.}
\label{overview_en_de}
\end{figure*}

The analyses mentioned above inspire us to design a novel non-unified training strategy for non-uniform deblurring. We propose a blurry component divided guided network which include two branches. One is a large blur reconstruction branch which is mainly used to constrain the processing of the large degree blurry regions recovery, and the other is a small blur reconstruction branch which pays more attentions on small degree blurry regions. These two different branches are restricted by the proposed adaptive component divided attention module (ACDA). The outputs of two branches are fed to an orientation-based feature fusion module which combines the large and small reconstructed features together and generates the finally deblurred image. The overview of our method is shown as Fig.\ref{overview_en_de}.
The framework of the proposed method consists of \textit{\textbf{feature extraction}}, \textit{\textbf{adaptive component divided attention}}, \textit{\textbf{two-branch reconstruction subnetwork}} and \textit{\textbf{orientation-based feature fusion}}.

\begin{equation}
\hat{I}=CDGNet(I_B,\theta),
\end{equation}
where $CDGNet(\cdot)$ represents the proposed deblur model, $I_B$ and $\hat{I}$ represent the input blurry image and deblurred image. $\theta$ is the parameter of the model.

\subsection{Feature Extraction}
\label{encoder_RDB}
Since deep residual network \cite{he2016deep}, DenseNet \cite{huang2017densely} and Residual DenseNet \cite{zhang2018residual} have been proposed, the recent encoder becomes more complex than only using the convolution layer, which can extract more abundant features of the input image. Compared to ResBlock and DenseBlock, the residual dense block (RDB) consists dense connected layers and local feature fusion with local residual learning, leading to a contiguous memory mechanism. 

\begin{figure*}[htbp]
\centering
\includegraphics[width=4in]{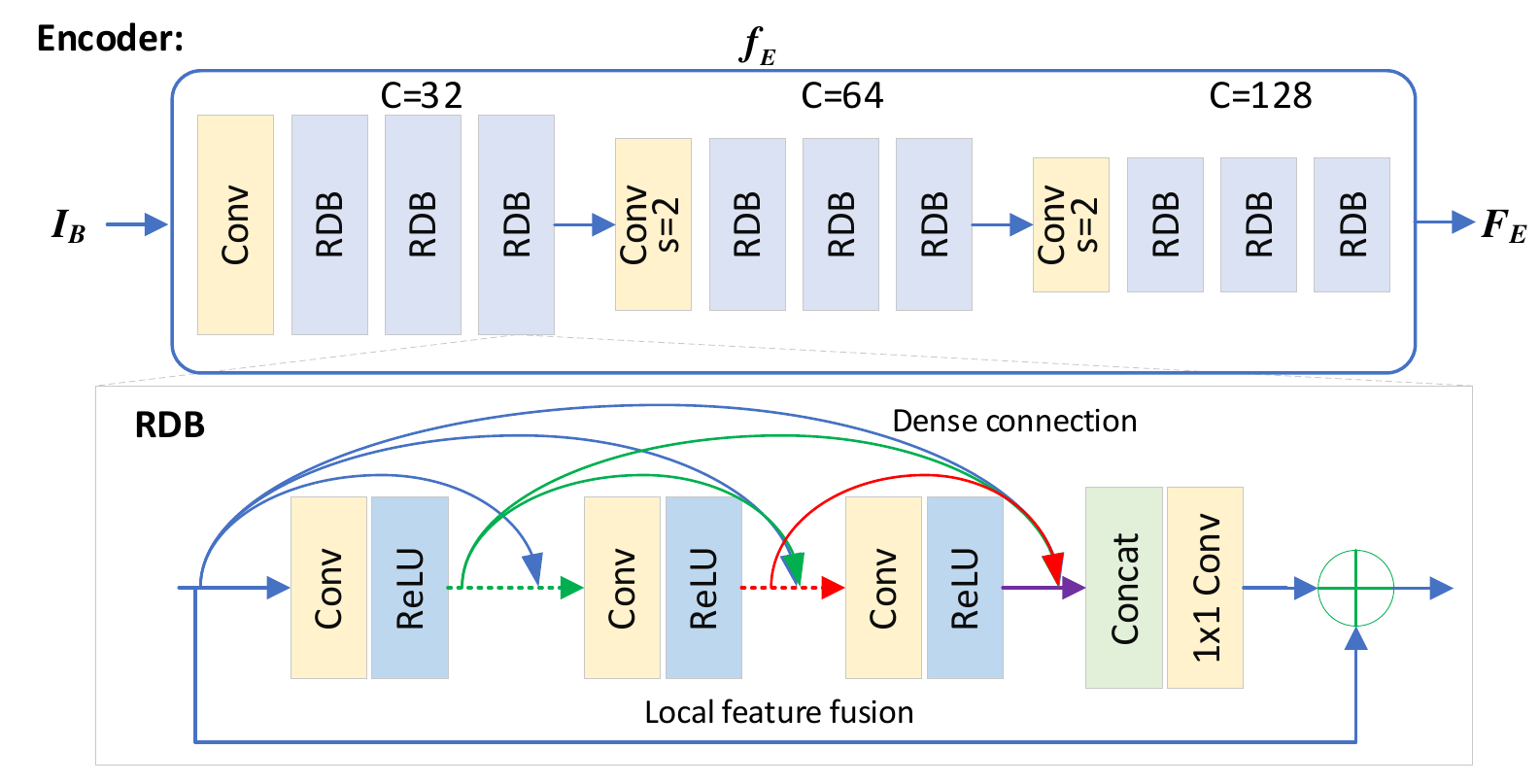}
\caption{The encoder's structure that includes several convolution layers and the RDB blocks.}
\label{RDB}
\end{figure*}

Drawing on their advantages, we design a RDB-based encoder with three resolution levels to achieve our feature extraction module $f_E$, which enlarges the feature receptive field by downsampling the input $I_B\in \mathbb{R}^{3\times H\times W}$ two times to help our model extract various level features $F_E\in \mathbb{R}^{C\times \frac{H}{4}\times \frac{W}{4}}$ for deblurring. $C$ is the number of channels, $C=128$ in our implementation. $H$ and $W$ are the height and width of the input. The structure of our encoder is shown in Fig.\ref{RDB}.
\begin{equation}
F_{E}=f_{E}(I_B).
\end{equation}

More specifically, our encoder consists of 3 convolution layers (the stride of the last two convolution layers for downsampling is 2) and 6 residual dense blocks (RDB). Each block includes 4 convolutional layers and 3 ReLU activation functions. 
As shown in Fig.\ref{overview_en_de}, the two branches employ the same encoder for an input blurry image.

\subsection{Adaptive Component Divided Attention}
\label{section_ACDA}
To divide the features into different regions according to the component of the blurry image, unlike the traditional encoder-decoder deblurring architecture, we specially design an attention module which named Adaptive Component Divided Attention module (ACDA). Recently, some attention mechanisms \cite{zhang2018image,woo2018cbam} simultaneously perform characteristic attention in feature's spatial and channel, and achieve remarkable results in computer vision task. Inspired by these methods, we develop a new adaptive component divided attention module based on CBAM, which changes the structure of the spatial and channel for better image deblurring. 

\begin{figure}[]
\centering
\includegraphics[width=4.2in]{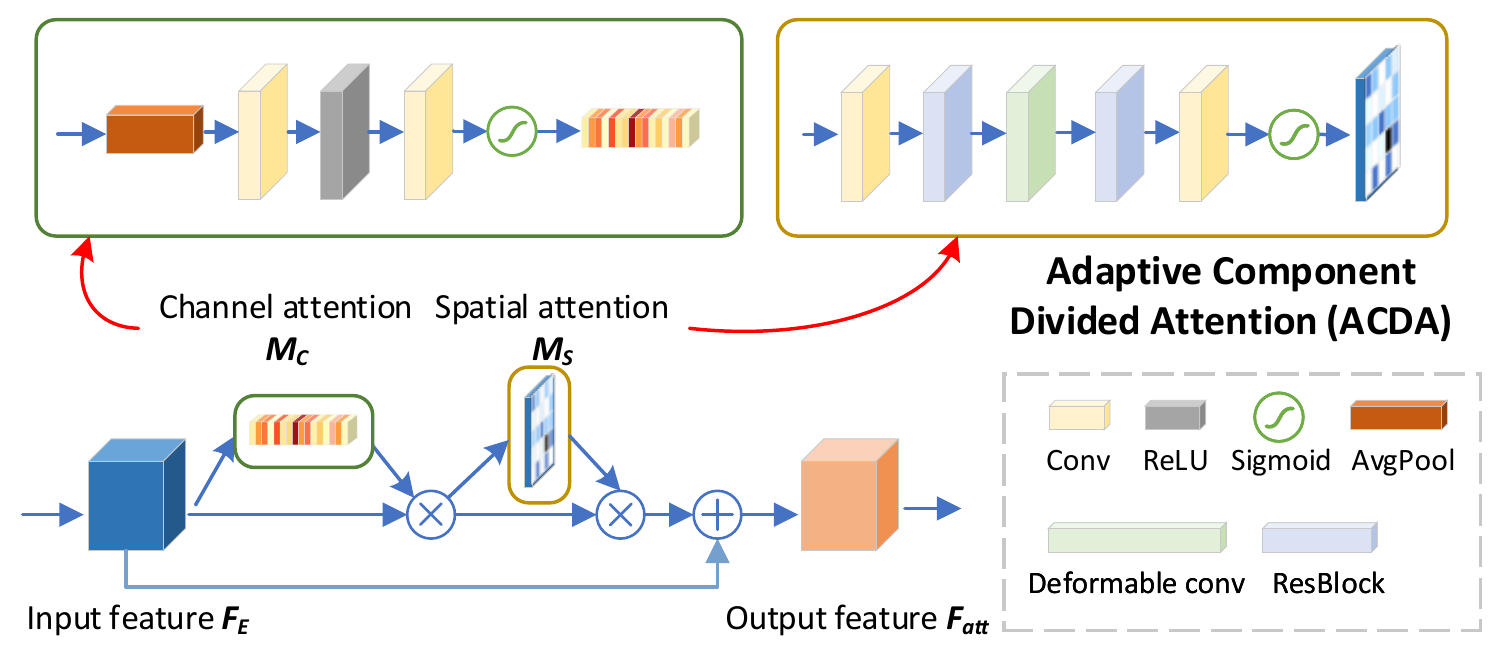}
\caption{The Details of our attention module which consists of the advanced channel and spatial attentions.}
\label{ACDA}
\end{figure}
As shown in Fig.\ref{ACDA}, our attention module is different from previous works. We improve the channel attention by deleting the MaxPooling layer since the max value seems useless to image restoration task \cite{zhang2018image}. For spatial attention, we apply several convolution layers, ResBlocks, deformable convolution and Sigmoid activation functions to build our spatial attention module which is proved effectively in subsequent experiments (see in Section \ref{Ablation}).

As shown in Fig.\ref{overview_en_de}, we employ two ACDA modules $f_{AL}$ and $f_{AS}$ that force the branches to pay more attention to their own regions: the large blurry regions and small blurry regions. In order to reduce the complexity of the model, we adopt the different parameters but the same structure of $f_{AL}$ and $f_{AS}$ to train the model. Take the large branch attention module as an example. Following the Section \ref{encoder_RDB}, 
$F_E\in \mathbb{R}^{C\times H'\times W'}$ denotes the input feature of ACDA, where $H'=\frac{H}{4}$, $W'=\frac{W}{4}$. Then the proposed attention module can be written as:

\begin{equation}
    \begin{aligned}
    F_{attL}&=f_{AL}(F_{E}) \\
            &= F_E\odot M_{c}\odot M_{s} + F_E,
    \end{aligned}
\end{equation}
where $F_{attL}\in \mathbb{R}^{C\times H'\times W'}$ represents the output of ACDA with the large blur branch. $\odot$ denotes element-wise multiplication. The channel attention map $M_{c} \in \mathbb{R}^{C\times 1\times 1}$ and spatial attention map $M_{s}\in \mathbb{R}^{1\times H'\times W'}$ are described in Eq.\ref{mc_ms},
where $\sigma$ represents the sigmoid functions,
$c1$ and $c2$ are a series of convolution layers, and $AvgPool$ is the global average pooling, whose implement can be found in Fig.\ref{ACDA}.
\begin{equation}
\label{mc_ms}
    \begin{aligned}
    M_{c}&=\sigma(c1(AvgPool(F_E))), \\
    M_{s}&=\sigma(c2(F_E\odot M_{c})).
    \end{aligned}
\end{equation}

Similarly, $F_{attS}\in \mathbb{R}^{C\times H'\times W'}$ represents the outputs of ACDA with the small blur branch.
\begin{equation}
    F_{attS}=f_{AS}(F_{E}).
\end{equation}

Based on the ACDA, we can obtain the large and small blur attention map. As shown in Fig.\ref{ACDA}, we conduct element-wise multiplication with the SA (spatial attention) maps and previous features, followed by adding the input feature. After feeding the ACDA's output features $F_{attL}$ and $F_{attS}$ to the two-branch decoders, we can finally divide the blurry regions into large blur and small blur.

\subsection{Two-branch Reconstruction Subnetwork}
\label{Two-branch Reconstruction Decoder Branch}
\begin{figure}[]
\centering
\includegraphics[width=4.5in]{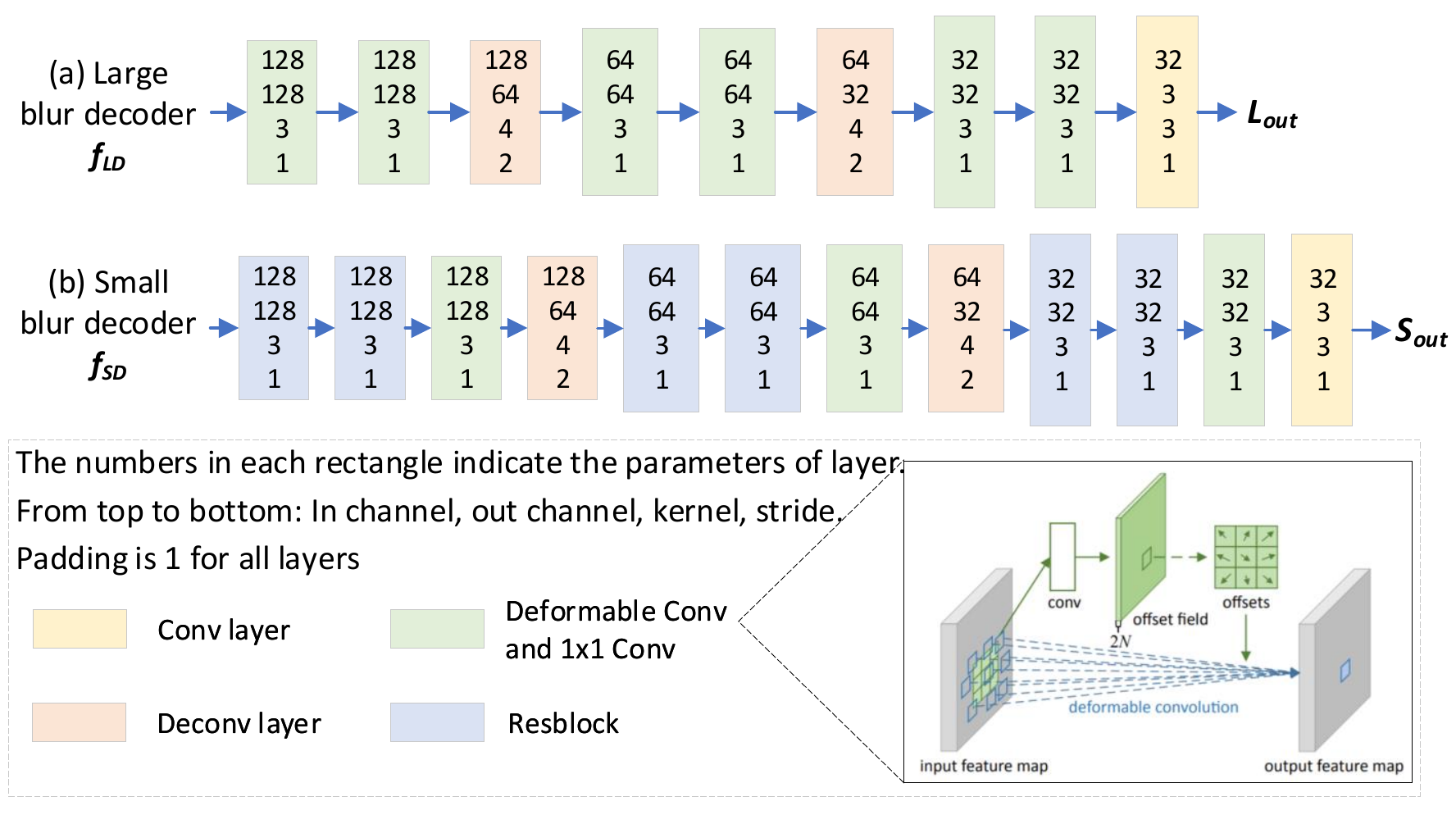}
\caption{Each layers of the large and small blur decoders module and the details of deformable convolution layer.}
\label{decoder}
\end{figure}

The illustrations in Section \ref{Introduction} have proved an important view that the large and small blur should be treated in different ways during the training stage. The deformable convolution can improve the capability of regular convolution and enlarge the feature receptive field. And its realization is related to the shape and position of the image which is consistent with the blur phenomenon. Thus, utilizing the deformable convolution to recover the latent sharp image is a practical method. Compared with the regular convolution operator, the deformable convolution in which additional offsets are learned to allow the network to obtain information away from its regular local neighborhood, improving the capability of regular convolutions. The details of the our decoders and the deformable convolution are shown in Fig.\ref{decoder}.

In this way, we propose two reconstruction decoders including several deformable convolution layers to recover the sharp features. Because of the huge gap between the blurry region and the corresponding sharp region, and the larger the blur, the harder the gap to handle. Thus, our large blur decoder is constructed by more layers of  deformable convolution than the small blur branch, which aims to deal with the large blurry regions.

The large blur decoder is shown in Fig.\ref{decoder}(a). We replace the common ResBlock \cite{zhang2019deep} with deformable convolution layers to recover the sharp regions corresponding to large blur. Considering that the deformable convolution layers can enlarge the receptive field, this branch can remove the large blur of the edges, and can constrain the branch to learn the optimal parameters that covered a wide area of the image structure. The large blur decoder $f_{LD}$ reconstruct the features $L_{out}\in \mathbb{R}^{C''\times H\times W}$ of the sharp regions according to the following equations: 
\begin{equation}
    L_{out} = f_{LD}(F_{attL}),
\end{equation}
where $F_{attL}$ is the large ACDA's output, $C''=32$. As well as the small blur decoder $f_{SD}$, where $S_{out}\in \mathbb{R}^{C''\times H\times W}$ is the large blur decoder's output.
\begin{equation}
    S_{out} = f_{SD}(F_{attS}).
\end{equation}

Fig.\ref{decoder}(b) shows each layer of the small blur branch decoder. This decoder adds an additional deformable convolutional layer at each feature level based on \cite{zhang2019deep}'s structure, which outperforms in the details and flat areas of blurry image that are difficult to recover with other algorithms. Each deformable convolution layer is followed by a $1\times1$ convolutional layer to reduce the feature dimension.

\subsection{Orientation-based Feature Fusion}
As mentioned above, the large blur branch mainly deals with the strong edge's blur, and the small blur branch mainly processes the detail and flat area's blur. Different features extracted from the same image always reflect the different characteristics of images. By optimizing and combining these different features, it not only keeps the effective discriminant information of multi-feature, but also eliminates the redundant information to certain degree \cite{sun2005new}. The feature maps after the two branches have the characteristics of their respective regions: the strong edges of the large blurry area are clear, and the details of the flat area are enhanced obviously. 
\begin{figure}[]
\centering
\includegraphics[width=3.5in]{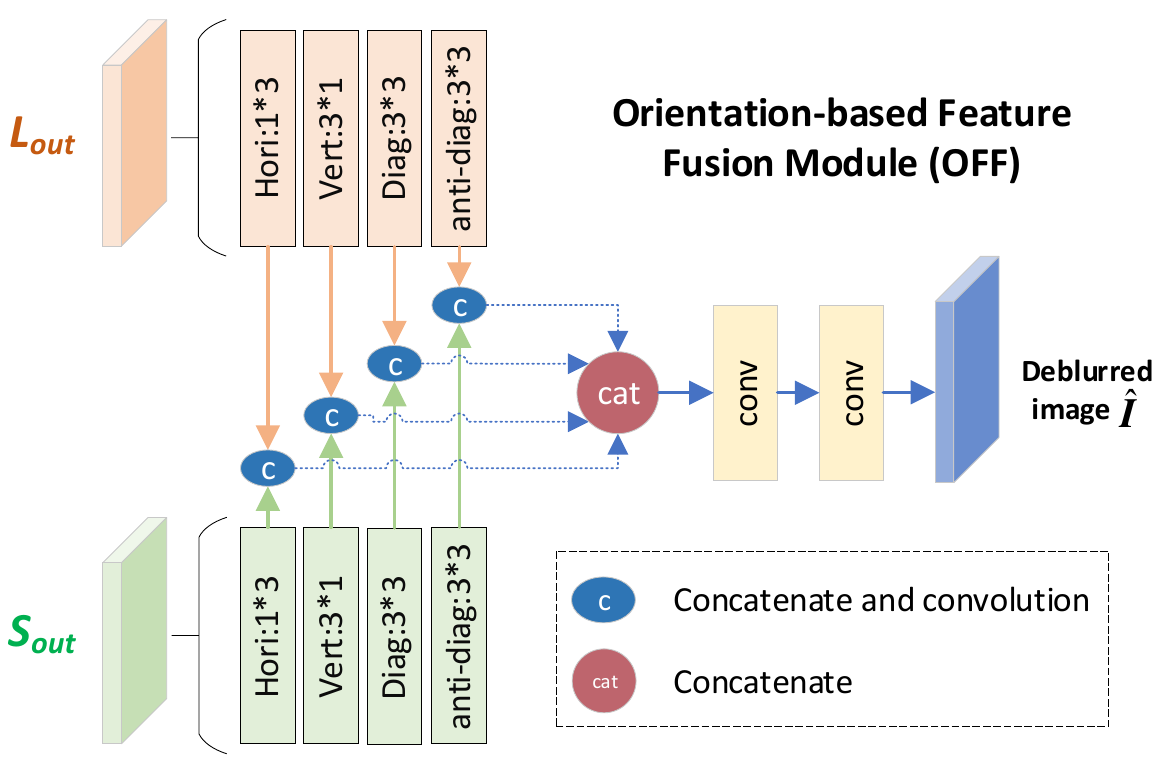}
\caption{Our orientation-based fusion module which consists of four orientation: horizontal, vertical, main-diagonal and anti-diagonal.}
\label{fusion}
\end{figure}

In order to combine the large and small reconstructed features $L_{out}$ and $S_{out}$, we propose an orientation-based feature fusion module (OFF) that relies on feature orientation filter. As shown below, $\hat{I}$ is the final deblurred image. 
\begin{equation}
    \hat{I}=OFF(L_{out},S_{out}).
\end{equation}

The structure of this module is shown in Fig.\ref{fusion}. Our orientation-based fusion module consists of four orientations: horizontal $H_{o}$ which is built by $1\times3$ kernel convolution, vertical $V_{o}$ with $3\times1$ kernel, main-diagonal $D_{o}$ and anti-diagonal $antiD_{o}$ are implemented by $3\times3$ kernel.
\begin{equation}
    \begin{aligned}
     L_{hori} &= H_{o}(L_{out}), 
     \:\:\:  L_{vert}= V_{o}(L_{out}), \\
     L_{diag} &= D_{o}(L_{out}),
     \:\:\:  L_{adiag} = antiD_{o}(L_{out}).
     \end{aligned}
\end{equation}

After applying the same operator on the small decoder output $S_{out}$, we can obtain the $S_{hori}$, $S_{vert}$, $S_{diag}$ and $S_{adiag}$. Then we combine the orientation-based features:
\begin{equation}
    \begin{aligned}
     F_{hori} &=Conv(cat(L_{hori},S_{hori})), 
     \:\:\:  F_{vert} =Conv(cat(L_{vert},S_{vert})), \\
     F_{diag} &=Conv(cat(L_{diag},S_{diag})), 
     \:\:\:  F_{adiag} =Conv(cat(L_{adiag},S_{adiag})).
     \end{aligned}
\end{equation}

Finally, the deblurred image $\hat{I}$ can be produced by merging the each orientation's features:
\begin{equation}
   \hat{I} =Conv(cat(F_{hori},F_{vert},F_{diag},F_{adiag})),
\end{equation}
where $cat$ indicates the concatenation which is shown in the Fig.\ref{fusion}.

\subsection{Non-unified Training Strategy and Loss Function}
\label{loss}
During the training stage, based on the above designed model, we propose a non-unified training strategy which has three loss functions to constrain the model to learn the optimal parameters: reconstruction loss, large blur reconstruction loss and small blur reconstruction loss. Relying on the two-branch decoder structures and additional loss functions, our model can be trained in a non-unified manner which aims to remove the large and small blurry regions, respectively. 

\textbf{Reconstruction loss.} Similar to other image generate tasks, our network also trains the $L_2$ loss as our model's reconstruction loss. Although some models adopted the other losses as their training objective functions, such as $L_1$ loss and SSIM loss \cite{zhao2016loss}, in Section \ref{Ablation}, we will demonstrate that the $L_2$ loss is the best choice for our model. The loss equation can be written as:
\begin{equation}
\label{re_loss}
\mathcal{L}_{rec}=\frac{1}{N}\sum_{i=1}^{N}\left \| \hat{I}(x_i) - I_{gt}(x_i) \right \|_{2}^{2},
\end{equation}
where $\hat{I}$ represents the output of our model, $I_{gt}$ is the ground truth corresponding to the input image, $N$ is the pixel number of the image, and the $\left \| \cdot\right \|_{2}^{2} $ is L2 norm.

\textbf{Large and small blur reconstruction loss.} As we mentioned above in Section \ref{section_ACDA} and \ref{Two-branch Reconstruction Decoder Branch}, our network relies on decomposition of different blurry regions, but how does our network become effective? In fact, during our training stage, we leverage a useful supervised signal, \textit{sharpness mask}, to help our model distinguish the blur. 

In specifically, we first use \cite{pan2020cascaded}'s method to obtain the each sharpness image of input blurry image, noted by $S$, which indicates the relative sharp region of input. Because our network divides the blurry image into two regions, we set $S$ as a binary image map. Then the \textit{sharpness mask} $\mathcal{M}$ can be acquired by the following equation: 
\begin{equation}
\label{eq6}
\mathcal{M}(x_{i})=max(0,sign(S(x_{i}) - \mu)),
\end{equation}
where $sign$ is signum function, $x_{i}$ is the value in pixel $i$. Based on the numerous experiments, we finally set the binary threshold $\mu$ as 0.96.

In this way we can obtain the important supervised signal $\mathcal{M}$. Further more, we multiply the signal mask $\mathcal{M}$ and ground truth $I_{gt}$ to produce the two-branch's ground truth images which are used to calculate the losses:
\begin{equation}
    \begin{aligned}
         S_{gt}(x_i) &= \mathcal{M}(x_i)\odot I_{gt}(x_i), \\
         L_{gt}(x_i) &= (1-\mathcal{M}(x_i))\odot I_{gt}(x_i),
    \end{aligned}
\end{equation}
where $S_{gt}$ and $L_{gt}$ are the ground truth of our small blur and large blur branch's output, respectively. Thus the loss function can be written as:
\begin{equation}
    \begin{aligned}
        \mathcal{L}_s&=\frac{1}{N}\sum_{i=1}^{N}\left \| S_{out}(x_i)  - S_{gt}(x_i)\right \|_{2}^{2},\\
        \mathcal{L}_l&=\frac{1}{N}\sum_{i=1}^{N}\left \| L_{out}(x_i)  - L_{gt}(x_i)\right \|_{2}^{2},
    \end{aligned}
\end{equation}
where $L_{s}$ and $L_{l}$ are small and large blur reconstruction loss, respectively. $S_{out}$ and $L_{out}$ represent the output of the small deblur branch and large deblur branch.

\textbf{Total loss.} 
In order to drive the network to learn the optimal parameters both on large and small blur regions, we set different weights for the three losses and obtain the total loss.
\begin{equation}
loss = \mathcal{L}_{rec} + \lambda_ 1\ast \mathcal{L}_s + \lambda_ 2\ast \mathcal{L}_l,
\end{equation}
where $\lambda_ 1$ and $\lambda_ 2$ are all 0.1 at training stage for better results. We also attempted the others values of $\lambda$ which can be seen in Section \ref{Ablation}.

\section{Experiments}
\label{Experiments}
\subsection{Training settings and implementation details} 
All our experiments are implemented in PyTorch and evaluated on a single NVIDIA RTX 1080Ti GPU. To train our network, we randomly crop input images to $256\times256$ pixel size. The batch size is set to 6 during training. The Adam solver is used to train our models for 3000 epochs. The initial learning rate is set to 0.0001, the decay rate set to 0.5 and step size is 500. We normalize image to range [0,1] and then subtract 0.5, so that our input's range is [-0.5,0.5]. 

\subsection{Datasets and Compared Methods} 
\subsubsection{Datasets}
{For fair comparison}, we train and test our {model} on the GoPro dataset \cite{nah2017deep}, then additionally test {it} on the VideoDeblurring dataset \cite{su2017deep} and some real-world blurry scenarios.

\textbf{GoPro dataset.} We train our network on dynamic scene deblurring benchmark GoPro dataset proposed by Nah \textit{et al.} \cite{nah2017deep}. This dataset is constructed from 240fps video captured by GoPro camera at $720\times1280$ resolution. The blurry images are generated by averaging varying number (7–13) of successive latent frames to produce varied blur. It contains 3214 pairs images where 2103 pairs were for training and 1111 pairs for testing.

\textbf{VideoDeblurring dataset}  is a video deblurring dataset \cite{su2017deep} which consists 71 videos. Every video consists of 100 frames at $720\times1280$ resolution. Likes GoPro dataset, they also synthesize blurry images {by} averaging high frame rate images. The difference to GoPro dataset is that it contains videos captured by various devices, such as iPhone, GoPro and Nexus.

\textbf{Real-world dataset.} 
{It} is difficult to capture both blur and clear images of the same scene at the same time. Thus, the above mentioned datasets are all synthesized, which may still differ from real blurry images. Therefore, we evaluate the performance of our network in some real scenarios. Our real-world blurry images are from the Pascal VOC \cite{everingham2010pascal} and RWBI \cite{zhang2020deblurring} dataset. They are all captured in natural scenes with real motion blur by varying hand-held devices directly. 

\subsubsection{Compared Methods}
We compare our method with previous state-of-the-art methods for image deblurring and carry out quantitative and qualitative comparisons of our architectures on both evaluation datasets and real images. Since our model aims to deal with non-uniform blur, it's unfair to compare with traditional uniform deblurring methods. {The compared methods} are all deep-learning based method and deal with dynamic blurring majority, include Sun 2015 \cite{sun2015learning}, Gong 2017 \cite{gong2017motion}, MS-CNN \cite{nah2017deep}, SRN \cite{tao2018scale-recurrent}, SVRNN \cite{zhang2018dynamic}, DMPHN \cite{zhang2019deep} and DeblurGAN-v2 \cite{kupyn2019deblurgan}, etc. Public implementations with default parameters were used to obtain qualitative results on test images. Some methods' code are unavailable, we use the results in their paper.

In recent two years, some methods achieved the highest performance for deblurring that regard compute-intensive structures as their crucial module. Purohit \textit{et al.} \cite{purohit2020region} leveraged the ImageNet pre-trained DenseNet as the encoder. Suin \textit{et al.} \cite{suin2020spatially} adopted very complicate connection links with each modules and proposes several compute blocks. Sim \textit{et al.} \cite{sim2019deep} used the adjacent frames of input as the extra temporal information to help the task that also made great progress. Their model size should be larger than ours which only uses a few residual blocks and deformable convolutions. Thus, for fair comparisons, we don't compare with those methods in this paper.

\subsection{Experimental Results}
\subsubsection{Quantitative Evaluation}

In order to {conduct} a fair comparison, same as other methods, our network is only trained on GoPro dataset \cite{nah2017deep}, and test on GoPro and VideoDeblurring \cite{su2017deep} to evaluate our model's extensibility and  robustness. The results are listed in Table \ref{GoPro_restable} and \ref{DVD_restable}. 
\begin{table*}[]
\setlength{\belowcaptionskip}{5pt}
\caption{Quantitative results on \textit{GoPro} dataset. 
Parameter size is in MB. FLOPs is in Gmac of $1280\times720$ images. Running time is in milliseconds. Best and second best results are \textbf{highlighted} and \underline{underlined}. '---' represents their code are not available.}
\label{GoPro_restable}
\centering
\scalebox{0.8}{
\begin{tabular}{l|c|c|c|c|c|c}
\hline
\textbf{Methods} &\textbf{Training sets} & \textbf{PSNR$_{\uparrow}$} & \textbf{SSIM$_{\uparrow}$} & \textbf{Param$_{\downarrow}$} & \textbf{FLOPs$_{\downarrow}$} & \textbf{Time$_{\downarrow}$} \\
\hline
Sun 2015 \cite{sun2015learning}  			& \cite{everingham2010pascal}   & 24.64 & 0.843 & 54.1 & {---} & {---} \\
Gong 2017 \cite{gong2017motion} 			& \cite{arbelaez2010contour,lin2014microsoft}     	& 26.06 & 0.863 & 41.2 &{---} & {---} \\
\hline
MS-CNN 2017 \cite{nah2017deep}  			& \cite{nah2017deep}	  & 26.97 & 0.839 & 303.6 & 1760.04 & 2400 \\
SRN 2018 \cite{tao2018scale-recurrent}   		& \cite{nah2017deep}    	& 30.19 & 0.938 & 33.6  &1434.82 & 1293 \\
SVRNN 2019 \cite{zhang2018dynamic}   		& \cite{nah2017deep,su2017deep}	 	& 29.19 & 0.931 & 37.1  & {---}   & {---} \\
DeblurGAN-v2 2019 \cite{kupyn2019deblurgan} 	& \cite{nah2017deep,su2017deep,Galoogahi2017need}	   & 28.03 & 0.848  & 243.7  & {\underline{411.34}} & 171  \\
HAMD 2019 \cite{ShenWLSLX019}  			& \cite{nah2017deep,ShenWLSLX019}   & {30.26} & \underline{0.940} & {---} & {---} & {---}\\
DMPHN 2019 \cite{zhang2019deep}	& \cite{nah2017deep,su2017deep}   & {{30.21}} & 0.898 & {\textbf{21.7}} & 689.20 & {\underline{7}} \\
DBGAN 2020 \cite{zhang2020deblurring}	 		& \cite{nah2017deep,zhang2020deblurring}    & \underline{30.43} & 0.937 & {---} & {---} & {---} \\
SVDN 2020 \cite{yuan2020efficient}	 		& \cite{nah2017deep,yuan2020efficient}    & 29.81 & 0.937 & --- & {---} & --- \\
\hline
\textbf{Ours} 							& \cite{nah2017deep} 	& \textbf{30.67} & {\textbf{0.945}} & \underline{25.57} &\textbf{292.57} &  \textbf{6} \\
\hline
\end{tabular}
}
\end{table*}

The average PSNR and SSIM measure obtained on the GoPro test split is provided in Table \ref{GoPro_restable}. It can be observed from the quantitative measures that our method performs better compared to previous state-of-the-art methods. We also evaluate the model's FLOPs, size and running time. As reported in Table \ref{GoPro_restable}, the results of our method get comparable PSNR (30.67), but higher SSIM (0.945), lower model size (25.57M) and FLOPs ({292.57G}) compared to other methods, {thanks} to our simple model structure and the framework which can deal with large and small blur {together}. The results shown in last row of Table \ref{GoPro_restable} whose value is the highest with lower computational cost. Additionally, we list the training sets of each method (the 2nd column of Table \ref{GoPro_restable}). Our models only are trained in GoPro dataset but achieve the extraordinary performance compared with other methods that using many training datasets.

\begin{table}[]
\setlength{\belowcaptionskip}{5pt}
\caption{Quantitative results on \textit{VideoDeblurring} dataset for models trained on \textit{GoPro} dataset, and finetune on the  \textit{VideoDeblurring} dataset.}
\label{DVD_restable}
\centering
\scalebox{0.8}{
\begin{tabular}{ccccc|c}
\toprule
\textbf{Methods}   & MS-CNN \cite{nah2017deep}   & SRN \cite{tao2018scale-recurrent} & DeblurGAN-v2 \cite{kupyn2019deblurgan}  & DMPHN\_1\_2\_4 \cite{zhang2019deep}   & \textbf{Ours}     \\ 
\midrule
\textbf{PSNR}      & 33.01   & {33.66}  &{33.91} & \underline{34.78}  & \textbf{35.25}  \\ 
\textbf{SSIM}      & 0.930   & 0.925 & 0.919         & \underline{0.966}   & \textbf{0.968}      \\ 
\bottomrule
\end{tabular}
}
\end{table}
We also evaluate our model on VideoDbelurring dataset. The results are shown in Table \ref{DVD_restable}. Our method also performs remarkable on the VideoDeblurring dataset, {especially for the highest SSIM (0.968)}, which means that our component divided guidance network not only can recover the latent sharp image, but also can maintain the image structural similarity effectively. 

\begin{figure*}[htbp]
\centering
\includegraphics[width=4.8in]{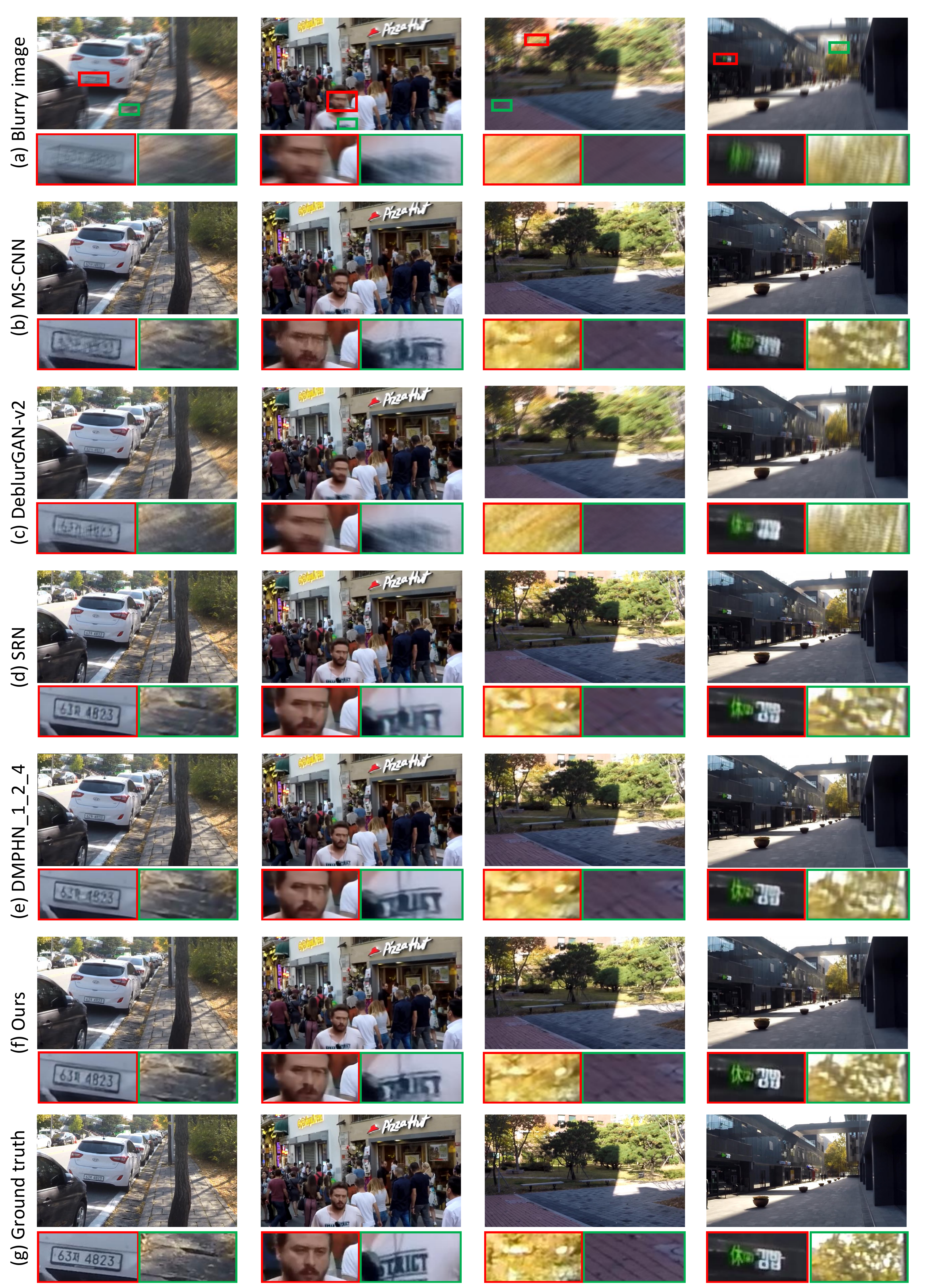}
\caption{Comparison with state-of-the-art deblurring methods on \textit{GoPro} dataset \cite{nah2017deep}. The improvement is clearly visible in the magnified patches, especially in the detail parts of our results have been greatly improved.}
\label{GoPro_res}
\end{figure*}

\subsubsection{Qualitative Evaluation}
Fig.\ref{GoPro_res} presents the final results of each method on the GoPro dataset, we also display the zoomed-in patches, which are marked as red and green boxes. 
We can find that the proposed method removes large and small blur and enhances the details of smooth areas, which proves the effectiveness of the two-branch strategy. For example, the words on the license plate on the first row the last column of Fig.\ref{GoPro_res} are very distinct, and each number can be identified, which is clearer than other methods.

\begin{figure}[htbp]
\centering
\includegraphics[width=4.6in]{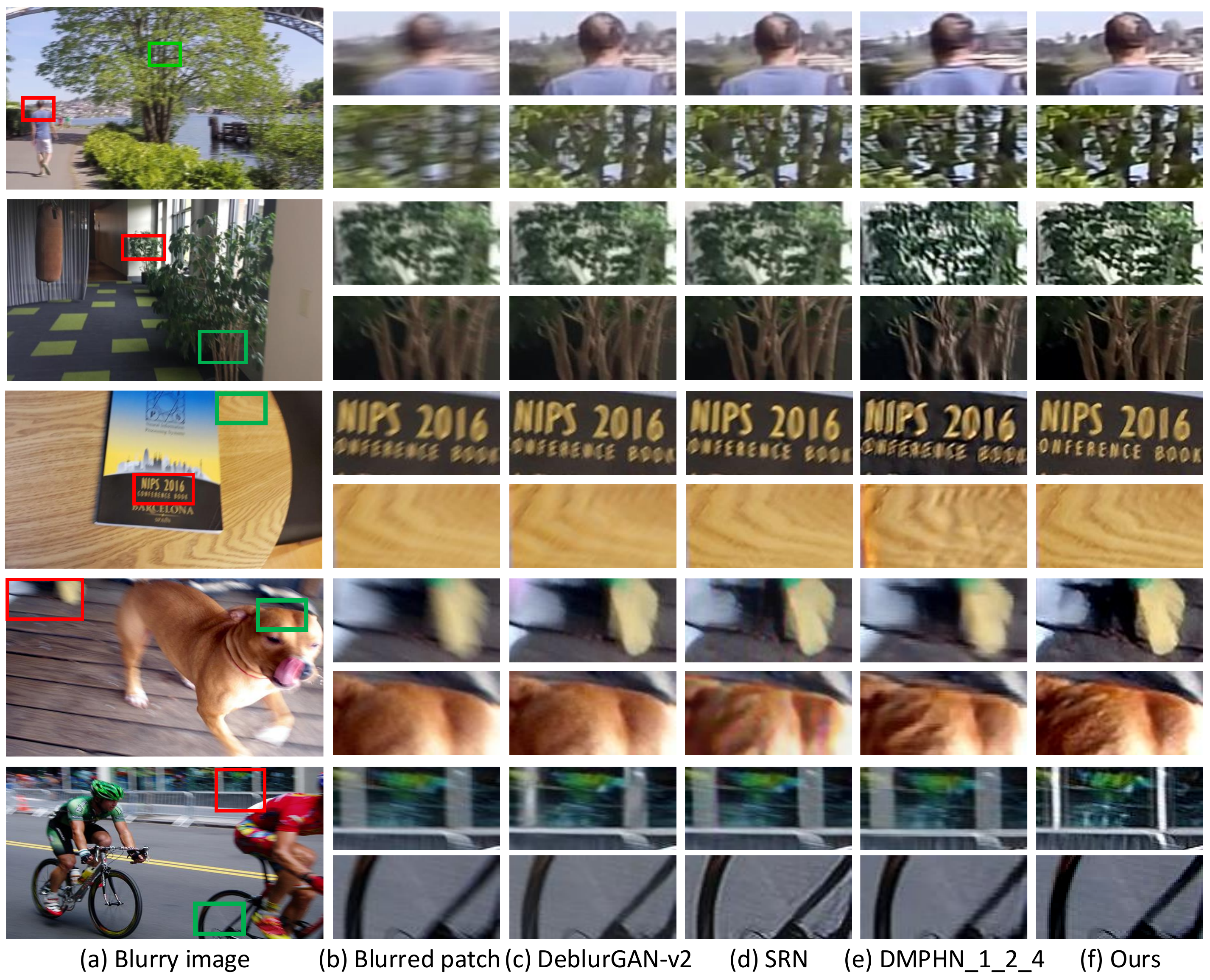}
\caption{Comparison with state-of-the-art deblurring methods on \textit{VideoDeblurring} dataset \cite{su2017deep} and real blurry images \cite{everingham2010pascal, zhang2020deblurring}. 
State-of-the-art methods generate the deblurred images with significant blur residual and artificial ringing, especially the leaves and the texts.}
\label{DVD_voc_res}
\end{figure}

In order to demonstrate the generalization of our model, Fig.\ref{DVD_voc_res} shows the deblurred results on \textit{VideoDeblurring} dataset \cite{su2017deep} and real-world blurry images (Pascal VOC \cite{everingham2010pascal} and RWBI \cite{zhang2020deblurring}). The results of state-of-the-art methods still exist the significant blur residual. Some artificial ringings appear in the results. In contrast, our method generates a much cleaner image, which is more similar to the sharp image. 

As shown in the last two rows of the Fig.\ref{DVD_voc_res}, our model can simultaneously handle the large blurry regions (the top row of each blurry image) and small blurry regions (the bottom row of each blurry image). The results are effective compared to SRN \cite{tao2018scale-recurrent} and DMPHN \cite{zhang2019deep} which are all SOTA methods. Our results enhance the texture and details, as well as remove the large blur.

\subsection{Ablation Studies}
\label{Ablation} 
In this section, we conduct some experiments to investigate the effectiveness of different modules of our model. Starting from the DMPHN\_1 \cite{zhang2019deep} which is our baseline model, we gradually inject our modifications on the structure. The results of adding or changing some modules based on DMPHN\_1 are shown in Table \ref{ablation_1}. 
\begin{table}[tbp]
\setlength{\belowcaptionskip}{10pt}
\caption{Ablation studies with different model on the \textit{GoPro} dataset.} 
\label{ablation_1}
\centering
\scalebox{0.7}{
\begin{tabular}{lccccccc}
\toprule
\textbf{Models}       & \textbf{Encoder}        	& \textbf{ACDA}       & \textbf{\begin{tabular}[c]{@{}c@{}}Large\\ decoder\end{tabular}}       &  \textbf{\begin{tabular}[c]{@{}c@{}}Small\\ decoder\end{tabular}}     &\textbf{Fusion} & \textbf{PSNR}           & \textbf{SSIM}  \\            	       \\ 
\hline
DMPHN\_1       	& ResBlock	&\xmark   &\xmark      &\xmark    &concatenate & 28.78   		&0.917     			\\ 
\hline
DMPHN+ACDA     & ResBlock   		& \cmark    &\xmark     &\xmark   &concatenate  & 28.79   		&0.918     			\\ 
DMPHN\_LD    	& ResBlock	&\xmark   &\cmark       &\xmark   &concatenate  & 28.88   		&0.920    			\\ 
Large blur branch  & ResBlock 		&\cmark    &\cmark       &\xmark   &concatenate  & 29.34   		&0.929     			\\ 
DMPHN\_SD    & ResBlock   		&\xmark  &\xmark     &\cmark      &concatenate   & 29.39    		&0.923     			\\ 
Small blur branch  & ResBlock  	&\cmark    &\xmark     &\cmark      &concatenate & 29.50   		&0.927     			\\ 
\hline
Two-branch+concat    & ResBlock     	&\cmark    &\cmark     &\cmark   &concatenate  & \textbf{30.05}& \textbf{0.939} 		\\ 
Two-branch+OFF & ResBlock &\cmark    &\cmark     &\cmark   &OFF  & \textbf{{30.24}}& \textbf{{0.941}} 		\\ 
RDB+Two-branch+OFF & RDB &\cmark    &\cmark     &\cmark   &OFF  & \textbf{30.67}& \textbf{0.945} 		\\ 
\bottomrule
\end{tabular}}
\end{table}

Firstly, we add an ACDA module between the DMPHN\_1's encoder and decoder, marking it as 'DMPHN+ACDA'. In order to illustrate the effectiveness of the designed decoder, we construct 'DMPHN\_LD' and 'DMPHN\_SD' that we replace original decoder of DMPHN with our large branch decoder and small branch decoder. Further, 'Large blur branch' and 'Small blur branch' demonstrate that the combing of the ACDA and large/small decoder outperforms the baseline. Finally, the two-branch models with the fusion module acquires the highest metrics in PSNR and SSIM. By the way, we also compare the results of different encoders and fusion modules of our model, which are listed in the last three rows of Table \ref{ablation_1}. 'Two-branch+concat' means that we use two branch decoders with ResBlocks' encoder and regard normal concatenation as our fusion module. After replacing the concatenation by proposed OFF, we can obtain the 'Two-branch+OFF'. Finally, 'RDB+Two-branch+OFF' is our full model which consists of the proposed encoder for feature extraction, ACDA for blurry regions division, two-branch subnetwork for reconstruction and OFF fusion module. It can be found that each module of our model is very important and boosts the model to achieve the best performance. Compared with our baseline model, DMPHN\_1, our full model improves 1.89/0.028 point in PSNR/SSIM with little model size increase. 

As one of core module, ACDA improves the performance obviously which can be found in Table \ref{ablation_ACDA}. We train a single large blur branch model with GoPro training datasets and test on its testing datasets. With the proposed ACDA, our model's results (PSNR/SSIM) improves dramatically from 28.88/0.920 to 29.34/0.929. {In the mean time}, when we only use channel attention (CA) or spatial attention (SA) module in our framework, our {method still shows favourable results} compared to \textit{without the attention} module.

\begin{table}[tbp]
\setlength{\belowcaptionskip}{10pt}
\caption{Ablation studies of different ACDA model. We train the \textit{single large blur branch} models with \textit{GoPro} training datasets and test on the testing datasets.}
\label{ablation_ACDA}
\centering
\scalebox{0.8}{
\begin{tabular}{cccccc}
\toprule
\textbf{Models}                      & \textbf{CA}        		& \textbf{SA}         		& \textbf{PSNR}           	& \textbf{SSIM}              	& \textbf{Size(MB)}                  \\ 
\midrule
w/o ACDA             &\xmark          &\xmark     	&28.88   		&0.920       			& 6.73          \\
only CA                     &\cmark         	&\xmark       	&29.28   		& 0.926     			& 6.74          \\
only SA                     &\xmark           &\cmark     	&29.30   		& 0.924     			& 7.66          \\
with ACDA                &\cmark          	&\cmark     	&\textbf{29.34}  &\textbf{0.929}     	& \textbf{7.67}           \\
\bottomrule
\end{tabular}}
\end{table}

\begin{figure}[htbp]
\centering
\includegraphics[width=4.7in]{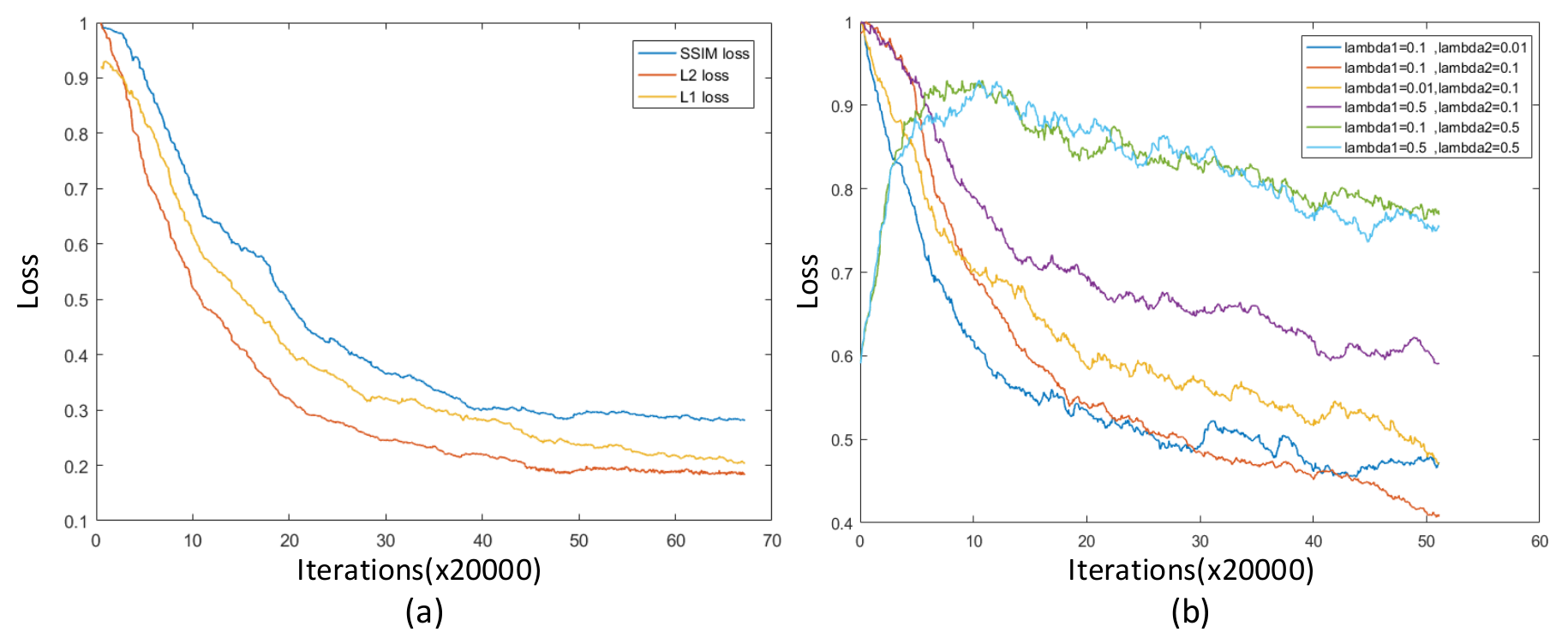}
\caption{The convergence curves of different loss functions and $\lambda$.}
\label{different_loss}
\end{figure}

\textbf{The effect of different loss functions.} In Section \ref{loss}, we have introduced our model that was trained with $L_2$ loss function. Actually, we also conducted experiments using different loss functions, the details are illustrated below. The $L_2$ loss for model's training achieves 30.67 PSNR and 0.945 SSIM. Using $L_1$ or $SSIM$ loss functions to constrain the network are becoming a trend in computer vision task, we also attempt these functions in our model. In contrast to other methods \cite{sim2019deep,park2020multi} which training with the $L_1$ and $SSIM$ loss, our model is converged to a score of 30.42/30.15 PSNR and 0.941/0.935 SSIM with $L_1$ and $SSIM$ loss. This is still considerably worse than our $L_2$ loss model (30.67/0.945). The convergence curves with different loss functions are shown in Fig.\ref{different_loss}(a). It demonstrated that the $L_2$ loss is more suitable to our model.
\begin{figure}[htbp]
\centering
\includegraphics[width=4.5in]{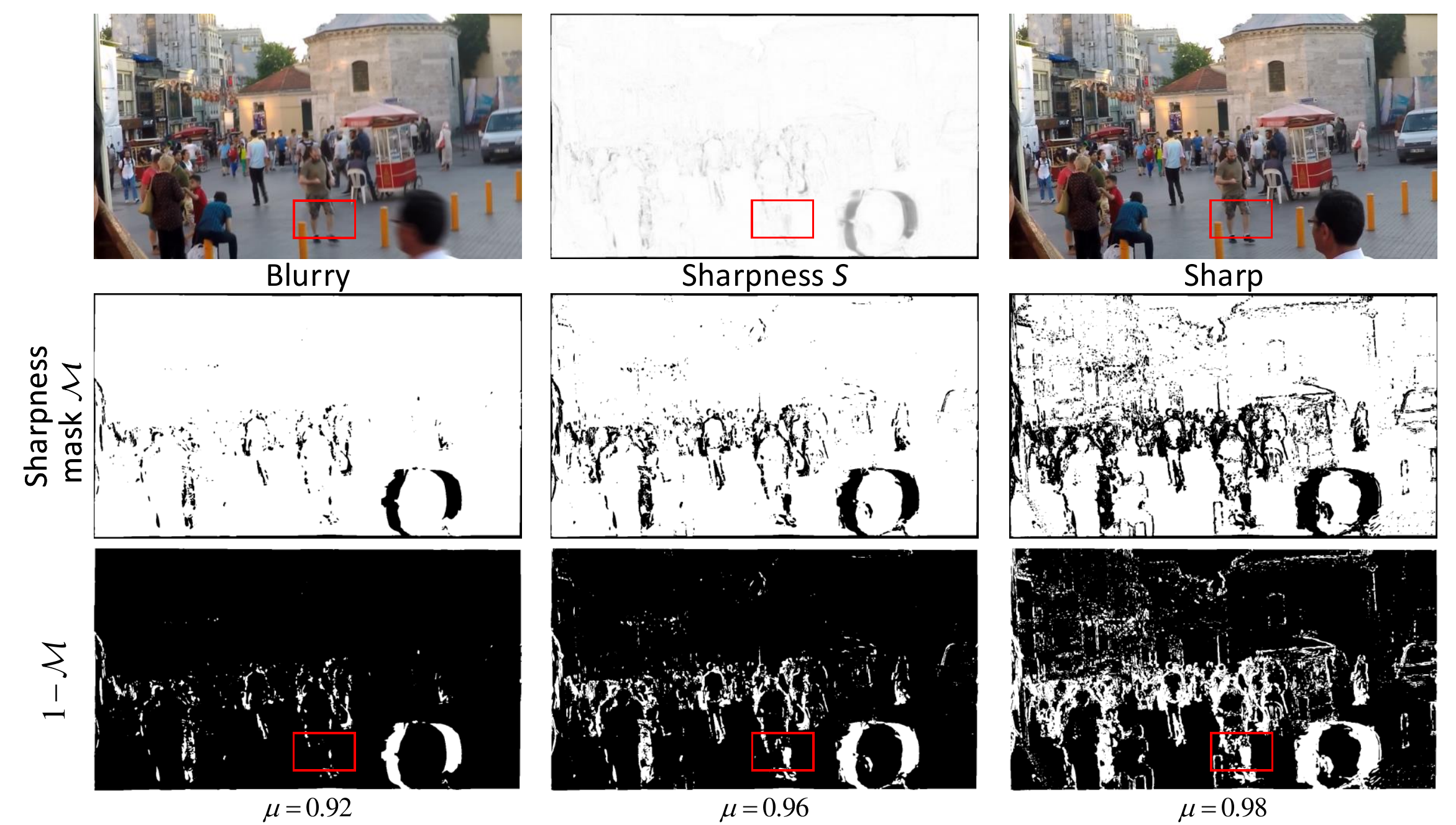}
\caption{The different $\mu$ resulting in different sharpness mask $\mathcal{M}$.}
\label{sharpness}
\end{figure}

Additionally, we carry out many experiments with the different $\mu$ and $\lambda$ to explain why we set $\mu=0.96$ and $\lambda_1=\lambda_2=0.1$ in our loss function. As the Section.\ref{loss} mentioned, $\mu$ is the threshold of the sharpness mask $\mathcal{M}$ which distinguish the large and small blurry regions of the sharpness $S$, and the $\lambda$ is related to the weight of each loss in the total loss. As shown in Fig.\ref{different_loss}(b) that is the convergence curve of the different $\lambda$ in our training stage, we can observe that the $\lambda_1=\lambda_2=0.1$ (red line) is the best choice for our model because it has a lower value and still maintains a downward trend in the same training time. Fig.\ref{sharpness} has shown the different $\mu$ leading to different sharpness masks which can be seen that when $\mu=0.96$ the $\mathcal{M}$ is consistent with human's perception compared with the blurry and sharp images.

\begin{figure}[htbp]
\centering
\includegraphics[width=3.7in]{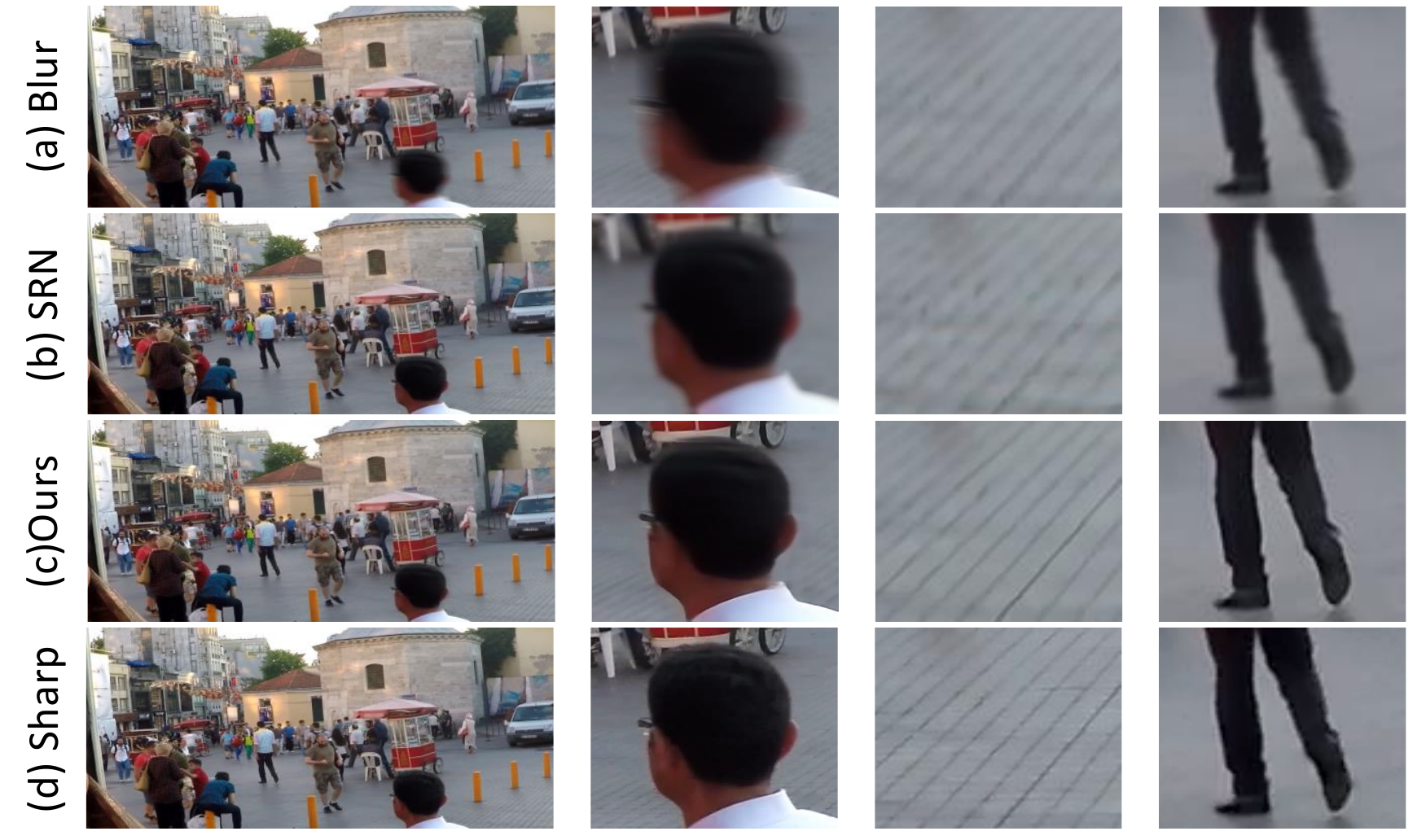}
\caption{The results of different regions of blurry image. Our model can remove large distinct blur (left patch), enhance the details (middle patch) and sharpen the smooth regions (right patch), simultaneously.}
\label{single_res}
\end{figure}

\textbf{Effectiveness of the Component Divided Guidance.} To demonstrate the effectiveness of the component divided guidance, we first crop three patches from a blurry image, as shown in Fig.\ref{single_res}, which shows the deblurring results of different blurry regions. From left to right, there are various results from one blurry image which validate that our proposed method is capable of removing large blur, reserving the details and sharpening the smooth regions, respectively.  

\begin{figure*}[htbp]
\centering
\includegraphics[width=4.8in]{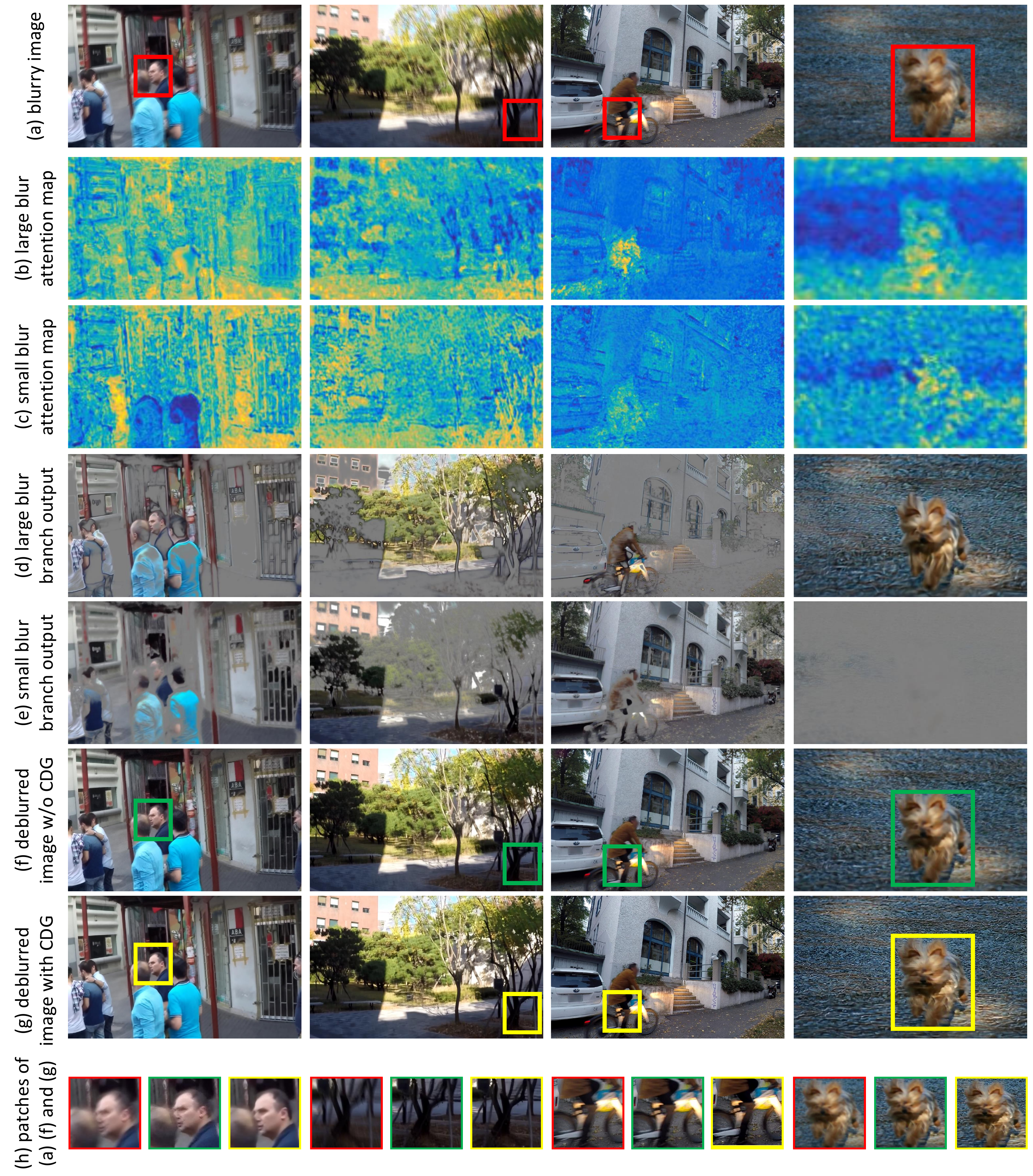}
\caption{Our attention maps and the deblurred results. From top to bottom: (a) represent blurry images, (b) and (c) are large and small blurry branch's attention maps, respectively. (d) and (e) are the outputs of the large blur branch and small blur branch. (f) and (g) are the deblurred results without and with the component divided guidance, respectively. Using the component divided guidance is able to generate much clearer images while the method without component divided guidance is less effective for blur removal.}
\label{attention_map_res}
\end{figure*}

Furthermore, Fig.\ref{attention_map_res} shows the visualization attention maps and the deblurred images of different component regions which are guided by the proposed component divided guidance network. Meanwhile, we compare the deblurred results without our component divided guidance (f) with our deblurred images with the guidance (g). As shown in the figures, our proposed method can deal with the non-uniform blurring favorably, where the large and small blur can be removed simultaneously and effectively.

\section{Conclusion}
\label{conclusion}
This paper {presents} a novel framework to deal with the non-uniform blur, which divides the blurry image via {the} adaptive component divided attention module (ACDA) with a two-branch training strategy. Our method makes full use of the correlation and complementary among different blurry regions. Subsequently, the image deblurring task can be implemented with additional orientation-based feature fusion module with different regions, which can further improve the deblurred image quality. In order to generate more effective attention map and deblurred images, three supervised reconstruction loss are employed to help the ACDA module to divide the large and small blurry regions. Extensive experiments show that the proposed algorithm based on this two-branch training manner converges quickly and can generate high-quality images with clear structures.

\section*{Acknowledgments}
This work is supported by the Project of the National Natural Science Foundation of China (Grant No.61901384 and No.61871328), as well as the Joint Funds of the National Natural Science Foundation of China (Grant No.U19B2037).

\bibliography{mybibfile.bib}

\begin{thebibliography}{10}
\expandafter\ifx\csname url\endcsname\relax
  \def\url#1{\texttt{#1}}\fi
\expandafter\ifx\csname urlprefix\endcsname\relax\def\urlprefix{URL }\fi
\expandafter\ifx\csname href\endcsname\relax
  \def\href#1#2{#2} \def\path#1{#1}\fi

\bibitem{Xu2010Two}
L.~Xu, J.~Jia, Two-phase kernel estimation for robust motion deblurring, in:
  Computer Vision - ECCV 2010, European Conference on Computer Vision,
  Heraklion, Crete, Greece, September 5-11, 2010, Proceedings, 2010, pp.
  157--170.

\bibitem{Cho2009Fast}
S.~Cho, S.~Lee, Fast motion deblurring, Acm Transactions on Graphics 28~(5)
  (2009) 1--8.

\bibitem{gong2017mpgl}
D.~Gong, M.~Tan, Y.~Zhang, A.~van~den Hengel, Q.~Shi, Mpgl: An efficient
  matching pursuit method for generalized lasso, in: AAAI Conference on
  Artificial Intelligence, 2017.

\bibitem{hsieh109blind}
P.~W. Hsieh, P.~C. Shao, Blind image deblurring based on the sparsity of patch
  minimum information, Pattern Recognition 109  107597.

\bibitem{peng2020joint}
J.~Peng, Y.~Shao, N.~Sang, C.~Gao, Joint image deblurring and matching with
  feature-based sparse representation prior, Pattern Recognition (2020) 107300.

\bibitem{li2019new}
T.~Li, H.~Chen, M.~Zhang, S.~Liu, S.~Xia, X.~Cao, G.~S. Young, X.~Xu, A new
  design in iterative image deblurring for improved robustness and performance,
  Pattern recognition 90 (2019) 134--146.

\bibitem{levin2011understanding}
A.~Levin, Y.~Weiss, F.~Durand, W.~T. Freeman, Understanding blind deconvolution
  algorithms, IEEE transactions on pattern analysis and machine intelligence
  33~(12) (2011) 2354--2367.

\bibitem{pan2014deblurring}
J.~Pan, Z.~Hu, Z.~Su, M.-H. Yang, Deblurring text images via l0-regularized
  intensity and gradient prior, in: Proceedings of the IEEE Conference on
  Computer Vision and Pattern Recognition, 2014, pp. 2901--2908.

\bibitem{gong2016blind}
D.~Gong, M.~Tan, Y.~Zhang, A.~Van~den Hengel, Q.~Shi, Blind image deconvolution
  by automatic gradient activation, in: Proceedings of the IEEE Conference on
  Computer Vision and Pattern Recognition, 2016, pp. 1827--1836.

\bibitem{krishnan2009fast}
D.~Krishnan, R.~Fergus, Fast image deconvolution using hyper-laplacian priors,
  in: Advances in neural information processing systems, 2009, pp. 1033--1041.

\bibitem{sun2015learning}
J.~Sun, W.~Cao, Z.~Xu, J.~Ponce, Learning a convolutional neural network for
  non-uniform motion blur removal, in: Proceedings of the IEEE Conference on
  Computer Vision and Pattern Recognition, 2015, pp. 769--777.

\bibitem{nah2017deep}
S.~Nah, T.~Hyun~Kim, K.~Mu~Lee, Deep multi-scale convolutional neural network
  for dynamic scene deblurring, in: Proceedings of the IEEE Conference on
  Computer Vision and Pattern Recognition, 2017, pp. 3883--3891.

\bibitem{li2018learning}
X.~Li, M.~Liu, Y.~Ye, W.~Zuo, L.~Lin, R.~Yang, Learning warped guidance for
  blind face restoration, in: Proceedings of the European Conference on
  Computer Vision (ECCV), 2018, pp. 272--289.

\bibitem{tao2018scale-recurrent}
X.~Tao, H.~Gao, X.~Shen, J.~Wang, J.~Jia, Scale-recurrent network for deep
  image deblurring, in: Proceedings of the IEEE Conference on Computer Vision
  and Pattern Recognition, 2018, pp. 8174--8182.

\bibitem{zhang2019deep}
H.~Zhang, Y.~Dai, H.~Li, P.~Koniusz, Deep stacked hierarchical multi-patch
  network for image deblurring, in: Proceedings of the IEEE Conference on
  Computer Vision and Pattern Recognition, 2019, pp. 5978--5986.

\bibitem{suin2020spatially}
M.~Suin, K.~Purohit, A.~Rajagopalan, Spatially-attentive patch-hierarchical
  network for adaptive motion deblurring, in: Proceedings of the IEEE/CVF
  Conference on Computer Vision and Pattern Recognition, 2020, pp. 3606--3615.

\bibitem{purohit2020region}
K.~Purohit, A.~Rajagopalan, Region-adaptive dense network for efficient motion
  deblurring., in: AAAI, 2020, pp. 11882--11889.

\bibitem{kaufman2020deblurring}
A.~Kaufman, R.~Fattal, Deblurring using analysis-synthesis networks pair, in:
  Proceedings of the IEEE/CVF Conference on Computer Vision and Pattern
  Recognition, 2020, pp. 5811--5820.

\bibitem{Fergus2006}
R.~Fergus, B.~Singh, A.~Hertzmann, S.~T. Roweis, W.~T. Freeman, Removing camera
  shake from a single photograph, in: ACM SIGGRAPH 2006 Papers, SIGGRAPH '06,
  Association for Computing Machinery, 2006, p. 787–794.

\bibitem{chen2008robust}
J.~Chen, L.~Yuan, C.-K. Tang, L.~Quan, Robust dual motion deblurring, in: 2008
  IEEE Conference on Computer Vision and Pattern Recognition, IEEE, 2008, pp.
  1--8.

\bibitem{gong2017motion}
D.~Gong, J.~Yang, L.~Liu, Y.~Zhang, I.~Reid, C.~Shen, A.~Van Den~Hengel,
  Q.~Shi, From motion blur to motion flow: a deep learning solution for
  removing heterogeneous motion blur, in: Proceedings of the IEEE Conference on
  Computer Vision and Pattern Recognition, 2017, pp. 2319--2328.

\bibitem{gao2019dynamic}
H.~Gao, X.~Tao, X.~Shen, J.~Jia, Dynamic scene deblurring with parameter
  selective sharing and nested skip connections, in: Proceedings of the IEEE
  Conference on Computer Vision and Pattern Recognition, 2019, pp. 3848--3856.

\bibitem{kupyn2018deblurgan}
O.~Kupyn, V.~Budzan, M.~Mykhailych, D.~Mishkin, J.~Matas, Deblurgan: Blind
  motion deblurring using conditional adversarial networks, in: Proceedings of
  the IEEE conference on computer vision and pattern recognition, 2018, pp.
  8183--8192.

\bibitem{kupyn2019deblurgan}
O.~Kupyn, T.~Martyniuk, J.~Wu, Z.~Wang, Deblurgan-v2: Deblurring
  (orders-of-magnitude) faster and better, in: Proceedings of the IEEE
  International Conference on Computer Vision, 2019, pp. 8878--8887.

\bibitem{sim2019deep}
H.~Sim, M.~Kim, A deep motion deblurring network based on per-pixel adaptive
  kernels with residual down-up and up-down modules, in: Proceedings of the
  IEEE Conference on Computer Vision and Pattern Recognition Workshops, 2019.

\bibitem{Bahdanau2014}
D.~Bahdanau, K.~Cho, Y.~Bengio, Neural machine translation by jointly learning
  to align and translate, Vol. 1409, 2014.

\bibitem{Vaswani2017}
A.~Vaswani, N.~Shazeer, N.~Parmar, J.~Uszkoreit, L.~Jones, A.~N. Gomez,
  u.~Kaiser, I.~Polosukhin, Attention is all you need, in: Proceedings of the
  31st International Conference on Neural Information Processing Systems,
  NIPS'17, Curran Associates Inc., 2017, p. 6000–6010.

\bibitem{ChenXYT18}
X.~Chen, C.~Xu, X.~Yang, D.~Tao, Attention-gan for object transfiguration in
  wild images, European Conference on Computer Vision (ECCV) 11206 (2018)
  167--184.

\bibitem{MejjatiRTCK18}
Y.~A. Mejjati, C.~Richardt, J.~Tompkin, D.~Cosker, K.~I. Kim, Unsupervised
  attention-guided image-to-image translation, Annual Conference on Neural
  Information Processing Systems (NeurIPS) (2018) 3697--3707.

\bibitem{ShenWLSLX019}
Z.~Shen, W.~Wang, X.~Lu, J.~Shen, H.~Ling, T.~Xu, L.~Shao, Human-aware motion
  deblurring, International Conference on Computer Vision (ICCV) (2019)
  5571--5580.

\bibitem{zhang2018image}
Y.~Zhang, K.~Li, K.~Li, L.~Wang, B.~Zhong, Y.~Fu, Image super-resolution using
  very deep residual channel attention networks, in: Proceedings of the
  European Conference on Computer Vision (ECCV), 2018, pp. 286--301.

\bibitem{woo2018cbam}
S.~Woo, J.~Park, J.-Y. Lee, I.~So~Kweon, Cbam: Convolutional block attention
  module, in: Proceedings of the European conference on computer vision (ECCV),
  2018, pp. 3--19.

\bibitem{hu2019senet}
J.~Hu, L.~Shen, S.~Albanie, G.~Sun, E.~Wu, Squeeze-and-excitation networks,
  2019.
\newblock \href {http://arxiv.org/abs/1709.01507} {\path{arXiv:1709.01507}}.

\bibitem{Dai2017Deformable}
J.~Dai, H.~Qi, Y.~Xiong, Y.~Li, G.~Zhang, H.~Hu, Y.~Wei, Deformable
  convolutional networks, in: Proceedings of the IEEE international conference
  on computer vision, 2017, pp. 764--773.

\bibitem{WangCYDL19}
X.~Wang, K.~C.~K. Chan, K.~Yu, C.~Dong, C.~C. Loy, {EDVR:} video restoration
  with enhanced deformable convolutional networks, Conference on Computer
  Vision and Pattern Recognition Workshops (CVPRW) (2019) 1954--1963.

\bibitem{dogan2019exemplar}
B.~Dogan, S.~Gu, R.~Timofte, Exemplar guided face image super-resolution
  without facial landmarks, in: Proceedings of the IEEE Conference on Computer
  Vision and Pattern Recognition Workshops, 2019, pp. 0--0.

\bibitem{Li_2020_CVPR}
X.~Li, W.~Li, D.~Ren, H.~Zhang, M.~Wang, W.~Zuo, Enhanced blind face
  restoration with multi-exemplar images and adaptive spatial feature fusion,
  in: CVPR, 2020.

\bibitem{qiu2019embedded}
Y.~Qiu, R.~Wang, D.~Tao, J.~Cheng, Embedded block residual network: A recursive
  restoration model for single-image super-resolution, in: Proceedings of the
  IEEE International Conference on Computer Vision, 2019, pp. 4180--4189.

\bibitem{he2016deep}
K.~He, X.~Zhang, S.~Ren, J.~Sun, Deep residual learning for image recognition,
  in: Proceedings of the IEEE conference on computer vision and pattern
  recognition, 2016, pp. 770--778.

\bibitem{huang2017densely}
G.~Huang, Z.~Liu, L.~Van Der~Maaten, K.~Q. Weinberger, Densely connected
  convolutional networks, in: Proceedings of the IEEE conference on computer
  vision and pattern recognition, 2017, pp. 4700--4708.

\bibitem{zhang2018residual}
Y.~Zhang, Y.~Tian, Y.~Kong, B.~Zhong, Y.~Fu, Residual dense network for image
  super-resolution, in: Proceedings of the IEEE conference on computer vision
  and pattern recognition, 2018, pp. 2472--2481.

\bibitem{sun2005new}
Q.-S. Sun, S.-G. Zeng, Y.~Liu, P.-A. Heng, D.-S. Xia, A new method of feature
  fusion and its application in image recognition, Pattern Recognition 38~(12)
  (2005) 2437--2448.

\bibitem{zhao2016loss}
H.~Zhao, O.~Gallo, I.~Frosio, J.~Kautz, Loss functions for image restoration
  with neural networks, IEEE Transactions on computational imaging 3~(1) (2016)
  47--57.

\bibitem{pan2020cascaded}
J.~Pan, H.~Bai, J.~Tang, Cascaded deep video deblurring using temporal
  sharpness prior, in: Proceedings of the IEEE/CVF Conference on Computer
  Vision and Pattern Recognition, 2020, pp. 3043--3051.

\bibitem{su2017deep}
S.~Su, M.~Delbracio, J.~Wang, G.~Sapiro, W.~Heidrich, O.~Wang, Deep video
  deblurring for hand-held cameras, in: Proceedings of the IEEE Conference on
  Computer Vision and Pattern Recognition, 2017, pp. 1279--1288.

\bibitem{everingham2010pascal}
M.~Everingham, L.~Van~Gool, C.~K. Williams, J.~Winn, A.~Zisserman, The pascal
  visual object classes (voc) challenge, International journal of computer
  vision 88~(2) (2010) 303--338.

\bibitem{zhang2020deblurring}
K.~Zhang, W.~Luo, Y.~Zhong, L.~Ma, B.~Stenger, W.~Liu, H.~Li, Deblurring by
  realistic blurring, in: Proceedings of the IEEE/CVF Conference on Computer
  Vision and Pattern Recognition, 2020, pp. 2737--2746.

\bibitem{zhang2018dynamic}
J.~Zhang, J.~Pan, J.~Ren, Y.~Song, L.~Bao, R.~W. Lau, M.-H. Yang, Dynamic scene
  deblurring using spatially variant recurrent neural networks, in: Proceedings
  of the IEEE Conference on Computer Vision and Pattern Recognition, 2018, pp.
  2521--2529.

\bibitem{arbelaez2010contour}
P.~Arbelaez, M.~Maire, C.~Fowlkes, J.~Malik, Contour detection and hierarchical
  image segmentation, IEEE transactions on pattern analysis and machine
  intelligence 33~(5) (2010) 898--916.

\bibitem{lin2014microsoft}
T.-Y. Lin, M.~Maire, S.~Belongie, J.~Hays, P.~Perona, D.~Ramanan,
  P.~Doll{\'a}r, C.~L. Zitnick, Microsoft coco: Common objects in context, in:
  European conference on computer vision, Springer, 2014, pp. 740--755.

\bibitem{Galoogahi2017need}
H.~K. {Galoogahi}, A.~{Fagg}, C.~{Huang}, D.~{Ramanan}, S.~{Lucey}, Need for
  speed: A benchmark for higher frame rate object tracking, in: 2017 IEEE
  International Conference on Computer Vision (ICCV), 2017, pp. 1134--1143.

\bibitem{yuan2020efficient}
Y.~Yuan, W.~Su, D.~Ma, Efficient dynamic scene deblurring using spatially
  variant deconvolution network with optical flow guided training, in:
  Proceedings of the IEEE/CVF Conference on Computer Vision and Pattern
  Recognition, 2020, pp. 3555--3564.

\bibitem{park2020multi}
D.~Park, D.~U. Kang, J.~Kim, S.~Y. Chun, Multi-temporal recurrent neural
  networks for progressive non-uniform single image deblurring with incremental
  temporal training, in: European Conference on Computer Vision, Springer,
  2020, pp. 327--343.

\end{thebibliography}

\end{document}